\documentclass[twoside,11pt]{article}

\usepackage{jmlr2e}
\usepackage{times}
\usepackage{amsfonts}
\usepackage{amsmath}
\usepackage{amssymb}
\usepackage{latexsym}
\usepackage{color}
\usepackage{graphics}
\usepackage{enumerate}
\usepackage{amstext}
\usepackage{epsfig}

\def\Rset{\mathbb{R}}

\DeclareMathOperator*{\argmin}{\rm argmin}

\DeclareMathOperator{\Tr}{Tr}
\DeclareMathOperator{\range}{range}
\providecommand{\abs}[1]{\lvert#1\rvert}
\providecommand{\norm}[1]{\lVert#1\rVert}
\providecommand{\normK}[1]{{\lVert#1\rVert}_K}

\newcommand{\ipsfig}[2]{\scalebox{#1}{\psfig{#2}}}
\newcommand{\mat}[1]{{\mathbf #1}}
\newcommand{\set}[1]{\{#1\}}
\newcommand{\seq}[3]{\mathbf{#1}_{#2}^{#3}}
\newcommand{\partition}[3]{#1 = (#2, #3)}
\newcommand{\expect}[2]{\mathbb{E}_{#1}\left[ #2 \right]}

\newcommand{\ltwo}[1]{\norm{#1}_2}
\newcommand{\linf}[1]{\norm{#1}_{\infty}}
\newcommand{\iprod}[2]{\left\langle #1 , #2 \right\rangle}
\newcommand{\bloc}{\beta_{loc}}
\newcommand{\ignore}[1]{}

\newcommand{\Rad}{{\widetilde R}}
\providecommand{\normtr}[1]{\lVert#1\rVert}
\newcommand{\me}{\mat{e}}
\renewcommand{\u}{\mat{u}}
\renewcommand{\v}{\mat{v}}
\newcommand{\mh}{\mat{h}}
\newcommand{\y}{\mat{y}}
\newcommand{\I}{\mat{I}}
\newcommand{\K}{\mat{K}}
\renewcommand{\L}{\mat{L}}
\newcommand{\M}{\mat{M}}
\renewcommand{\P}{\mat{P}}

\newcommand{\bc}{\mathbf{c}}

\newcommand{\bv}{\mathbf{v}}
\newcommand{\bZ}{\mathbf{Z}}
\newcommand{\X}{{\cal X}}
\newcommand{\Y}{{\cal Y}}

\newcommand{\cV}{{\cal V}}

\newcommand{\e}{\epsilon}
\newcommand{\h}{\widehat}

\newcommand{\T}{\mathbf{T}}
\renewcommand{\H}{\mathbf{H}}
\newcommand{\LTR}{$\mathtt{LTR}$}
\newcommand{\CM}{\mathtt{CM}}

\newcommand{\LLR}{\mathtt{LL-Reg}}
\newcommand{\GMF}{\mathtt{GMF}}

\ShortHeadings{Stability Analysis of Transductive Regression
Algorithms}{Cortes, Mohri, Pechyony, and Rastogi} 
\firstpageno{1}

\begin{document}

\title{Stability Analysis and Learning Bounds for \\
Transductive Regression Algorithms}

\author{\name Corinna Cortes \email corinna@google.com \\
       \addr Google Research\\
       76 Ninth Avenue,\\
       New York, NY 10011, USA.
       \AND
       \name Mehryar Mohri \email mohri@cs.nyu.edu \\
       \addr Courant Institute of Mathematical Sciences  and Google Research\\
       251 Mercer Street,\\
       New York, NY 10012, USA.
       \AND
       \name Dmitry Pechyony\thanks{This author's work was done 
         at the Computer Science Department, Technion -
         Israel Institute of Technology.}
       \email 
       pechyony@nec-labs.com \\
       \addr NEC Laboratories America, \\
       4 Independence Way, Suite 200,\\
       Princeton, NJ 08648, USA. 
       \AND
       \name Ashish Rastogi \email rastogi@cs.nyu.edu \\
       \addr Google Research\\
       76 Ninth Avenue,\\
       New York, NY 10011, USA.}

\editor{X}

\maketitle

\begin{abstract}
  This paper uses the notion of algorithmic stability to derive novel
  generalization bounds for several families of transductive
  regression algorithms, both by using convexity and closed-form
  solutions. Our analysis helps compare the stability of these
  algorithms. \ignore{It suggests that several existing algorithms
    might not be stable but prescribes a technique to make them
    stable.}  It also shows that a number of widely used transductive
  regression algorithms are in fact unstable. Finally, it reports the
  results of experiments with local transductive regression
  demonstrating the benefit of our stability bounds for model
  selection, for one of the algorithms, in particular for determining
  the radius of the local neighborhood used by the algorithm.
\end{abstract}

\tableofcontents

\section{Introduction}

The problem of \emph{transductive inference} was originally
introduced by \cite{vapnik82}. Many learning problems in information
extraction, computational biology, natural language processing and
other domains can be formulated as a transductive inference
problem. In the transductive setting, the learning algorithm receives
both a labeled training set, as in the standard induction setting, and
a set of unlabeled test points. The objective is to predict the labels
of the test points. No other test points will ever be considered. This
setting arises in a variety of applications. Often, there are orders
of magnitude more unlabeled points than labeled ones and they have not
been assigned a label due to the prohibitive cost of labeling.  This
motivates the use of transductive algorithms which leverage the
unlabeled data during training to improve learning performance.

This paper deals with transductive regression, which arises in
problems such as predicting the real-valued labels of the nodes of a
fixed (known) graph in computational biology, or the scores associated
with known documents in information extraction or search engine tasks.
Several algorithms have been devised for the specific setting of
transductive regression
\citep{belkin-tr,chapelle,schuurmans,Cortes&Mohri2006}. Several other
algorithms introduced for transductive classification can be viewed in
fact as transductive regression ones as their objective function is
based on the square loss, for example, in
\cite{belkin04,belkin-tr}. \cite{Cortes&Mohri2006} gave explicit
VC-dimension generalization bounds for transductive regression that
hold for all bounded loss functions and coincide with the tight
classification bounds of \cite{vapnik98} when applied to
classification.

We present novel algorithm-dependent generalization bounds for
transductive regression. Since they are algorithm-specific, these
bounds can often be tighter than bounds based on general complexity
measures such as the VC-dimension. Our analysis is based on the notion
of algorithmic stability and our learning bounds generalize to the
transduction scenario the stability bounds given by
\cite{bousquet-jmlr} for the inductive setting and extend to
regression the stability-based transductive classification bounds of
\cite{dmitrycolt06}.

In Section~\ref{sec:preliminaries} we give a formal definition of the
transductive inference learning set-up, including a precise
description and discussion of two related transductive settings. We
also introduce the notions of cost and score stability used in the
following sections.

\begin{figure*}[t]
\begin{center}
\ipsfig{.5}{figure=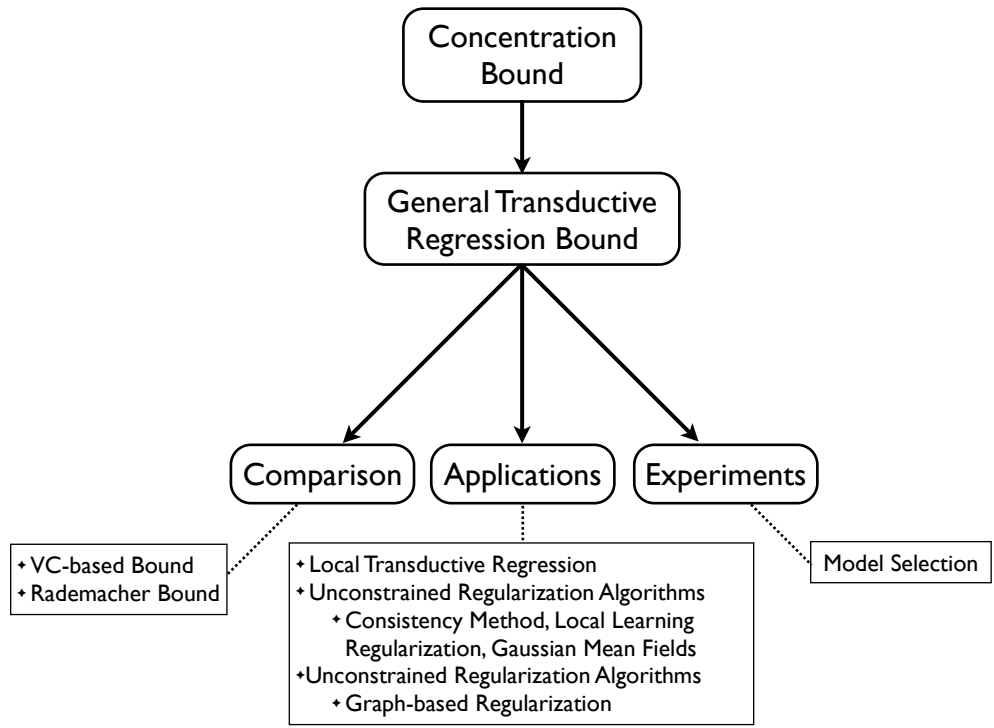}
\end{center}
\caption{A high-level outline of the paper.}
\label{fig:outline}
\end{figure*}

Standard concentration bounds such as McDiarmid's bound \citep{mcd89}
cannot be readily applied to the transductive regression setting since
the points are not drawn independently but uniformly without
replacement from a finite set. Instead,
Section~\ref{subsec:concentration bound} proves a concentration bound
generalizing McDiarmid's bound to the case of random variables sampled
without replacement. This bound is slightly stronger than that of
\cite{dmitrycolt06,dmitrycolt07} and the proof much simpler and more
concise. This concentration bound is used to derive a general
transductive regression stability bound in
Section~\ref{sec:transductive stability
  bound}. Figure~\ref{fig:outline} shows the outline of the paper.

Section~\ref{sec:stability of local transductive regression
  algorithms} introduces and examines a very general family of
tranductive algorithms, that of local transductive regression (\LTR)
algorithms, a generalization of the algorithm of
\cite{Cortes&Mohri2006}.  It gives general bounds for the stability
coefficients of \LTR\ algorithms and uses them to derive
stability-based learning bounds for these algorithms. The stability
analysis in this section is based on the notion of cost stability and
based on convexity arguments.

In Section~\ref{sec:unconstrained}, we analyze a general class of
unconstrained optimization algorithms that includes a number of recent
algorithms \citep{WuS07,ZhouBLWS03,ZhuGL03}. The optimization problems
for these algorithms admit a closed-form solution. We use that to give
a score-based stability analysis of these algorithms. Our analysis
shows that in general these algorithms may not be stable. In fact, in
Section~\ref{sec:lower_bound} we prove a lower bound on the stability
coefficient of these algorithms under some assumptions.

Section~\ref{sec:constrained} examines a class of constrained
regularization optimization algorithms for graphs that enjoy better
stability properties than the unconstrained ones just mentioned. This
includes the graph Laplacian algorithm of \cite{belkin04}.  In
Section~\ref{sec:constrained_score}, we give a score stability
analysis with novel generalization bounds for this algorithm, simpler
and more general than those given by
\cite{belkin04}. Section~\ref{sec:constrained_cost} shows that
algorithms based on constrained graph regularizations are in fact
special instances of the \LTR\ algorithms by showing that the
regularization term can be written in terms of a norm in a reproducing
kernel Hilbert space. This is used to derive a cost stability analysis
and novel learning bounds for the graph Laplacian algorithm of
\cite{belkin04} in terms of the second smallest eigenvalue of the
Laplacian and the diameter of the graph. Much of the results of these
sections generalize to other constrained regularization optimization
algorithms. These generalizations are briefly discussed in
Section~\ref{sec:general_case} where it is indicated, in particular,
how similar constraints can be imposed to the algorithms of
\cite{WuS07,ZhouBLWS03,ZhuGL03} to derive new and stable versions of
these algorithms.

Finally, Section~\ref{sec:experiments} shows the results of
experiments with local transductive regression demonstrating the
benefit of our stability bounds for model selection, in particular for
determining the radius of the local neighborhood used by the
algorithm, which provides a partial validation of our bounds and
analysis.

\section{Definitions}
\label{sec:preliminaries}

Let $\X$ denote the input space and $\Y$ a measurable subset of $\Rset$.

\subsection{Transductive learning set-up}

In transductive learning settings, the algorithm receives a labeled
training set $S$ of size $m$, $S \!=\! ((x_1, y_1), \ldots, (x_m,
y_m)) \!\in\! \X \times \Y$, and an unlabeled test set $T$ of size
$u$, $x_{m + 1}, \ldots, x_{m + u} \!\in\! X$. The transductive
learning problem consists of predicting accurately the labels $y_{m +
  1}, \ldots, y_{m + u}$ of the test examples, no other test example
is ever considered. Two different settings can be distinguished to
formalize this problem, see \citep{vapnik98}.

\paragraph{Setting 1}

In this setting, a full sample $X$ of $m + u$ examples is given. The
learning algorithm further receives the labels of a training sample
$S$ of size $m$ selected from $X$ uniformly at random without
replacement. The remaining $u$ unlabeled examples serve as a test
sample $T$. We denote by $\partition{X}{S}{T}$ a partitioning of $X$
into a training set $S$ and test set $T$.

\paragraph{Setting 2}

Here, the training sample $S$ and test sample $T$ are both drawn
i.i.d.\ according to some distribution $D$. The labeled sample $S$ and
the test points $T$, without their labels, are made available to the
learning algorithm.

As in previous theoretical studies of the transduction problem, e.g.,
\citep{vapnik98,meir,Cortes&Mohri2006,dmitrycolt06}, we analyze
setting~1 and derive generalization bounds for this specific setting.
However, as pointed out by \cite{vapnik98}, any generalization bound
in the setting we analyze directly yields a bound for setting~2
by taking the expectation.

The specific problem where the labels are real-valued numbers, as in
the case studied in this paper, is that of \emph{transductive
  regression}. It differs from the standard \emph{inductive
  regression} since the learning algorithm is given the unlabeled test
examples beforehand and can thus can possibly exploit that information
to improve its performance.

\subsection{Notions of stability}

We denote by $c(h, x)$ the cost of an error of a hypothesis $h$ on a
point $x$ labeled with $y(x)$. The cost function commonly used in
regression is the square loss $c(h, x) \!=\! [h(x) - y(x)]^2$.  We
shall assume a square loss for the remaining of this paper, but many
of our results generalize to other convex cost functions. The training
error $\h R(h)$ and test error $R(h)$ of a hypothesis $h$ are defined
as follows:
\begin{equation}
  \widehat{R}(h) = \frac{1}{m}
  \sum_{k=1}^m c(h, x_k) \qquad R(h) = \frac{1}{u} \sum_{k=1}^u c(h,
  x_{m+k}).
\end{equation}
The generalization bounds we derive are based on the notion of
algorithmic stability. We shall use the following two notions of
stability in our analysis.

\begin{definition}[Cost stability]
\label{defn:cost-beta-stability}
  Let $L$ be a transductive learning algorithm and let $h$ denote the
  hypothesis returned by $L$ for $\partition{X}{S}{T}$ and $h'$ the
  hypothesis returned for $\partition{X}{S'}{T'}$, where $S$ and $S'$
  differ in exactly one point. $L$ is said to be \emph{uniformly
    $\beta$-stable} with respect to the cost function $c$ if there
  exists $\beta \geq 0$ such that for all $x \in X$,
\begin{equation}
\big| c(h', x) - c(h, x) \big| \leq \beta.
\end{equation}
\end{definition}
\begin{definition}[Score stability]
\label{defn:score-beta-stability}
Let $L$ be a transductive learning algorithm and let $h$ denote the
hypothesis returned by $L$ for $\partition{X}{S}{T}$ and $h'$ the
hypothesis returned for $\partition{X}{S'}{T'}$. $L$ is said to be
\emph{uniformly $\beta$-stable} with respect to its output scores if
there exists $\beta \geq 0$ such that for all $x \in X$,
\begin{equation}
\big| h'(x) - h(x) \big| \leq \beta.
\end{equation}
\end{definition}
We will say that a hypothesis set $H$ is bounded by $B \!>\! 0$ when
$|h(x) - y(x)| \leq B$ for all $x \in X$ and $h \in H$. For such a
hypothesis set and the square loss, for any two hypotheses $h, h' \in
H$ and $x \in \X$, the following inequality holds:
\begin{align}
|c(h', x) - c(h, x)|
& = \big|\big[h'(x) - y(x)\big]^2 - \big[h(x) - y(x)\big]^2\big|\\
& = \big| h'(x) - h(x) \big| \big| h'(x) - y(x) + h(x) - y(x) \big|\\
& \leq 2B \, |h'(x) - h(x)|.
\end{align}
Thus, for $H$ bounded by $B$ and the square loss,
$\beta$-score-stability implies $2B\beta$-cost-stability.

For the remainder of this paper, unless otherwise specified, stability
is meant as cost-based stability.

\section{General transduction stability bounds}
\label{sec:transductive regression stability bounds}

Stability-based generalization bounds in the inductive setting are
derived using McDiarmid's inequality \citep{mcd89}.  The main
technique used is to show that under suitable conditions on the
stability of the algorithm, the difference of the test error and the
training error is sharply concentrated around its expected value, and
that this expected value itself is small. Roughly speaking, this
implies that with high probability, the test error is close to the
training error. Since the points in the training and test sample are
drawn in an i.i.d.~fashion, McDiarmid's inequality can be applied.

However, in the transductive setting, the sampling random variables
are not drawn independently. Thus, McDiarmid's concentration bound
cannot be readily used in this case. Instead, a generalization of
McDiarmid's bound that holds for random variables sampled without
replacement is needed. We present such a generalization in this
section with a concise proof. A slightly weaker version of this bound
with a somewhat more complex proof was derived by
\cite{dmitrycolt06,dmitrycolt07}.

\subsection{Concentration bound for sampling without replacement}
\label{subsec:concentration bound}

To derive this concentration bound, we use the method of averaged
bounded differences and the following theorem due to \cite{azuma} and
\cite{mcd89}.  where we denote by $\seq{S}{i}{j}$ the subsequence of
random variables $S_i, \ldots, S_j$ and write $\seq{S}{i}{j} =
\seq{x}{i}{j}$ as a shorthand for the event $S_i = x_i, \ldots, S_j =
x_j$.

\begin{theorem}[\cite{mcd89}, Th. 6.10]
\label{thm:mobd-mcd}
Let $\seq{S}{1}{m}$ be a sequence of random variables with each $S_i$
taking values in $\X$. Let $\phi\colon \X^m \to \Rset$ be a measurable
function satisfying the following conditions:
\begin{equation*}
  \forall i \in [1, m], \forall x_i, x'_i \in \X,
  \Big | \expect{\seq{S}{i+1}{m}}{\phi|\seq{S}{1}{i-1}, S_i = x_i} -
    \expect{\seq{S}{i+1}{m}}{\phi|\seq{S}{1}{i-1}, S_i = x'_i} \Big | \leq
  c_i.
\end{equation*}
Then, for all $\e \!>\! 0$,
\begin{equation}
  \Pr \Big[ \phi - \expect{}{\phi} \geq \e \Big] \leq \exp
  \bigg[ \frac{-2\e^2}{\sum_{i=1}^m c_i^2}\bigg].
\end{equation}
\end{theorem}
The following definition is needed for the presentation of our
concentration bound.
\begin{definition}[Symmetric Functions]
  A function $\phi\colon \X^m \to \Rset$ is said to be
  \emph{symmetric} if its value does not depend on the order of its
  arguments, that is for any two permutations $\sigma$ and $\sigma'$
  over $[1, m]$ and any $m$ points $x_1, \ldots, x_m \in \X$,
  $\Phi(x_{\sigma(1)}, \ldots, x_{\sigma(m)}) = \Phi(x_{\sigma'(1)},
  \ldots, x_{\sigma'(m)})$.
\end{definition}

\begin{theorem}[Concentration bound for sampling without replacement]
\label{theorem:main}
Let $\seq{x}{1}{m}$ be a sequence of random variables, sampled
uniformly without replacement from a fixed set $X$ of $m + u$
elements, and let $\phi\colon \X^m \to \Rset$ be a symmetric function
such that for all $i \in [1, m]$ and for all $x_1, \ldots, x_m \in X$
and $x'_1, \ldots, x'_m \in X$,
\begin{equation*}
\Big| \phi(x_1, \ldots, x_m) - \phi(x_1, \ldots, x_{i-1}, x'_i,
x_{i+1}, \ldots, x_m)\Big| \leq c.
\end{equation*}
Then, for all $\e \!>\! 0$,
\begin{equation}
  \label{eqn:main-eqn}
  \Pr \Big[ \phi - \expect{}{\phi} \geq \e \Big]
  \leq  \exp \bigg[ \frac{-2 \e^2}{\alpha (m, u) c^2} \bigg],
\end{equation} 
where $\alpha (m, u) = \frac{mu}{m + u - 1/2} \, \frac{1}{1 - 1 /
(2\max\{m, u\})}$.
\end{theorem}

\begin{proof}
  Fix $i \in [1, m]$ and define $g(\seq{S}{1}{i-1}, x_i, x'_i)$ as
  follow:
\begin{equation} 
g(\seq{S}{1}{i-1}, x_i, x'_i) = \expect{\seq{S}{i+1}{m}}{\phi | \seq{S}{1}{i-1},
S_i = x_i} - \expect{\seq{S}{i+1}{m}}{\phi | \seq{S}{1}{i-1}, S_i =
x'_i}.
\end{equation}
Then,
\begin{align*}
  g(\seq{x}{1}{i-1}, x_i, x'_i) = & \sum_{\seq{x}{i+1}{m}} \phi(\seq{x}{1}{i-1},
  x_i, \seq{x}{i+1}{m}) \Pr[ \seq{S}{i+1}{m} = \seq{x}{i+1}{m} |
  \seq{S}{1}{i-1} = \seq{x}{1}{i-1}, S_i = x_i ] \nonumber \\
  & - \sum_{\seq{x'}{i+1}{m}} \phi(\seq{x}{1}{i-1}, x'_i,
  \seq{x'}{i+1}{m}) \Pr[ \seq{S}{i+1}{m} = \seq{x'}{i+1}{m} |
  \seq{S}{1}{i-1} = \seq{x}{1}{i-1}, S_i = x'_i ].
\end{align*}
We show that $g(\seq{x}{1}{i-1}, x_i, x'_i)$ can be bounded by $c_i =
\frac{uc}{m + u - i}$ and apply Theorem~\ref{thm:mobd-mcd} to obtain
the bound claimed. For uniform sampling without replacement, the
probability terms can be written explicitly:
\begin{equation*}
\Pr \left[ \seq{S}{i+1}{m} = \seq{x}{i+1}{m} | \seq{S}{1}{i-1} =
    \seq{x}{1}{i-1}, S_i = x_i \right] = \prod_{k = i}^{m - 1}
    \frac{1}{m + u - k} = \frac{u!}{(m + u - i)!}.
\end{equation*}
Thus,
\begin{equation*}
  g(\seq{x}{1}{i-1}, x_i, x'_i) = \frac{u!}{(m + u - i)!} \Bigl[
  \sum_{\seq{x}{i+1}{m}} \phi(\seq{x}{1}{i-1}, x_i, \seq{x}{i+1}{m}) -
  \sum_{\seq{x'}{i+1}{m}} \phi(\seq{x}{1}{i-1}, x'_i, \seq{x'}{i+1}{m})
  \Bigr].
\end{equation*}
To compute $\sum_{\seq{x}{i+1}{m}} \phi(\seq{x}{1}{i-1}, x_i,
\seq{x}{i+1}{m}) - \sum_{\seq{x'}{i+1}{m}} \phi(\seq{x}{1}{i-1}, x'_i,
\seq{x'}{i+1}{m})$, we divide the set of permutations $\{
\seq{x'}{i+1}{m} \}$ into two sets, those that contain the element
$x_i$ and those that do not. If a permutation $\seq{x'}{i+1}{m}$
contains $x_i$ we can write it as
$\seq{x'}{i+1}{k-1}x_i\seq{x'}{k+1}{m}$, where $k$ is such that $x'_k
= x_i$. We then match it up with the permutation $x_i
\seq{x'}{i+1}{k-1}\seq{x'}{k+1}{m}$ from the set $\{x_i
\seq{x}{i+1}{m}\}$. These two permutations contain exactly the same
elements, and since the function $\phi$ is symmetric in its arguments,
the difference in the value of the function on the permutations is
zero.

In the other case, if a permutation $\seq{x'}{i+1}{m}$ does not
contain the element $x_i$, then we simply match it up with the same
permutation in $\{\seq{x}{i+1}{m}\}$. The matching permutations
appearing in the summation are then $x_i \seq{x'}{i+1}{m}$ and $x'_i
\seq{x'}{i+1}{m}$ which clearly only differ with respect to $x_i$. The
difference in the value of the function $\phi$ in this case can be
bounded by $c$. The number of such permutations can be counted as
follows: it is the number of permutations of length $m - i$ from the
set $X$ of $m + u$ elements that do not contain any of the elements of
$\seq{x}{1}{i-1}$, $x_i$ and $x'_i$, which is equal to $\frac{(m + u -
  i - 1)!}{(u - 1)!}$. This leads us to the following upper bound on
$\sum_{\seq{x}{i+1}{m}} \phi(\seq{x}{1}{i-1}, x_i, \seq{x}{i+1}{m}) -
\sum_{\seq{x'}{i+1}{m}} \phi(\seq{x}{1}{i-1}, x'_i,
\seq{x'}{i+1}{m})$:
\begin{equation}
  \sum_{\seq{x}{i+1}{m}} \phi(\seq{x}{1}{i-1}, x_i,
  \seq{x}{i+1}{m}) - \sum_{\seq{x'}{i+1}{m}} \phi(\seq{x}{1}{i-1},
  x'_i, \seq{x'}{i+1}{m}) \leq \frac{(m + u - i - 1)!}{(u - 1)!} c,
\end{equation}
which implies that $ |g(\seq{x}{1}{i-1}, x_i, x'_i)| \leq \frac{u!}{(m
  + u - i)!}  \cdot \frac{(m + u - i - 1)!c}{(u - 1)!}  \leq
\frac{uc}{m + u - i} $.  To apply Theorem~\ref{thm:mobd-mcd}, we need
to bound $\sum_{i=1}^m \big( \frac{uc}{m + u - i} \big)^2$. To this
end, note that
\begin{equation}
\sum_{i=1}^m \frac{1}{(m + u - i)^2} = \sum_{j=u}^{m + u - 1}
\frac{1}{j^2} \leq \int_{u-1/2}^{m + u - 1/2} \frac{dx}{x^2} =
\frac{m}{m + u - 1/2} \, \frac{1}{u - 1/2}.
\end{equation}
The application of Theorem~\ref{thm:mobd-mcd} then yields:
\begin{equation}
\Pr \Big[ \phi -  \expect{}{\phi}  \geq \e \Big]
\leq \exp \bigg[ \frac{-2 \e^2}{\alpha_u (m, u) c^2} \bigg],
\end{equation}
where $\alpha_u (m, u) = \frac{mu}{m + u - 1/2} \, \frac{1}{1 - 1 /
  (2u)}$. Function $\phi$ is symmetric in $m$ and $u$ in the sense
that selecting one of the sets uniquely determines the other. The
statement of the theorem then follows by obtaining a similar bound
with $\alpha_m (m, u) = \frac{mu}{m + u - 1/2} \, \frac{1}{1 - 1 /
  (2m)}$ and taking the tighter of the two bounds.
\end{proof}

\subsection{Transductive stability bound}
\label{sec:transductive stability bound}

Observe that, since the full sample $X$ is given, the average error of
a hypothesis $h \!\in\! H$ over $X$ defined by $R_X(h) = \frac{1}{m +
  u}\sum_{i = 1}^{m + u} h(x_i)$ is not a random variable. Also, for
any training sample $S$, the test error $R(h)$ can be expressed in
terms of $R_X(h)$ and the empirical error $\h R(h)$ as follows:
\begin{equation}
R(h)  = \frac{1}{u} \sum_{i = 1}^{u} h(x_{m + i}) = \frac{1}{u} \Big((m + u) R_X(h) - \sum_{i = 1}^{m} h(x_i)\Big) = \frac{m + u}{u} R_X(h) - \frac{m}{u} \h R(h).
\end{equation}
Thus, for a fixed $h$, the quantity $R(h) - \h R (h) = \frac{m + u}{u}
R_X(h) - \frac{m + u}{u} \h R(h)$ only varies with $\h R(h)$ and is
only a function of the training sample $S$. Let $\Phi$ be defined by
$\phi (S) = R(h) - \h R (h)$. Since permuting the points of $S$ does
not affect $\h R(h)$, $\Phi$ is symmetric.

To obtain a general transductive regression stability bound, we apply
the concentration bound of Theorem~\ref{theorem:main} to the random
variable $\phi (S)$. To do so, we need to bound $\expect{S}{\phi(S)}$,
where $S$ is a random subset of $X$ of size $m$, and $|\phi(S) -
\phi(S')|$ where $S$ and $S'$ are samples differing by exactly one
point. The following lemma proves a Lipschitz condition for $\Phi$.

\begin{lemma}
\label{lemma:diff(phi)}
Let $H$ be a hypothesis set bounded by $B$. Let $L$ be a
$\beta$-cost-stable algorithm and let $S$ and $S'$ be two training
sets of size $m$ that differ in exactly one point. Let $h \in H$ be
the hypothesis returned by $L$ when trained on $S$ and $h' \in H$ the
one returned when $L$ is trained on $S'$. Then,
\begin{equation} 
|\phi(S) - \phi (S')| \leq 2\beta + \frac{m + u}{mu} B^2.
\end{equation} 
\end{lemma}
\begin{proof}
  By the definition of $S'$, there exist $i \in [1, m]$ and $j \in [1,
  u]$ such that $S' = S \setminus \{ x_i \} \cup \{ x_{m+j}
  \}$. $\phi(S) - \phi(S')$ can written as follows:
\begin{eqnarray*}
\phi(S) - \phi(S') & = & \frac{1}{u} \sum_{k=1, k\neq j}^u \left[c(h,
x_{m+k}) - c(h', x_{m+k})\right] + \frac{1}{m} \sum_{k=1, k\neq i}^m \left[c(h',
x_k) - c(h, x_k)\right] \\ & & + \frac{1}{u} \left[c(h, x_{m+j}) - c(h', x_{i})\right] +
\frac{1}{m} \left[c(h', x_{m+j}) - c(h, x_i)\right].
\end{eqnarray*}
Since the hypothesis set $H$ is bounded by $B$, the square loss $c$
is bounded by $B^2$, $c(h, x) \leq B^2$ for all $x \in X, h \in
H$. Thus,
\begin{equation} |\phi(S) - \phi(S')| \leq \frac{(u - 1) \beta}{u}  +
\frac{(m - 1)\beta}{m} + \frac{B^2}{u} + \frac{B^2}{m} \leq 2\beta +
B^2 \left( \frac{1}{u} + \frac{1}{m} \right).
\end{equation}
\end{proof}
The next lemma bounds the expectation of $\Phi$.

\begin{lemma}
\label{lemma:expect(phi)}
Let $h$ be the hypothesis returned by a $\beta$-cost-stable algorithm
$L$. Then, the following inequality holds for the expectation of $\Phi$:
\begin{equation}
\big| \expect{S}{\phi(S)} \big| \leq \beta.
\end{equation}
\end{lemma}

\begin{proof}
By the definition of $\phi(S)$, we can write
\begin{equation}
    \expect{S}{\phi(S)} = \expect{S}{R(h)} -
    \expect{S}{\h R(h)} =  \frac{1}{u} \sum_{k=1}^u
    \expect{S}{c(h, x_{m+k})} - \frac{1}{m} \sum_{k=1}^m
    \expect{S}{c(h, x_k)}.
\end{equation}
$\expect{S}{c(h, x_{m+k})}$ is the same for all $1 \leq k \leq u$, and
similarly, $\expect{S}{c(h, x_k)}$ is the same for all $1 \leq k \leq
m$.  Let $i \in [1, m]$ and $j \in [1, u]$, and let $S'$ be defined as
in the previous lemma: $S' = S \setminus \set{x_i} \cup
\set{x_{m+j}}$, and let $h'$ denote a hypothesis trained on $S'$, then
the following holds:
\begin{align}
\expect{S}{\phi(S)}
& =  \expect{S}{c(h, x_{m+j})} - \expect{S}{c(h, x_{i})}\\
& = \expect{S'}{c(h', x_i)} - \expect{S}{c(h, x_i)}\\
& = \expect{S, S'}{c (h', x_i) - c(h, x_i)} \leq \beta,
\end{align}
by the cost $\beta$-stability of the algorithm.
\end{proof}

\begin{theorem}
  \label{thm:generalization} Let $H$ be a hypothesis set bounded by
  $B$ and $L$ a $\beta$-cost-stable algorithm. Let $h$ be the
  hypothesis returned by $L$ when trained on
  $\partition{X}{S}{T}$. Then, for any $\delta \!>\! 0$, with
  probability at least $1 - \delta$,
\begin{equation}
\label{eqn:main_stability_bound}
R(h) \leq \h R(h) + \beta + \bigg(2\beta +
  \frac{B^2 (m + u)}{mu} \bigg) \sqrt{\frac{ \alpha (m, u) \ln {1
      \over \delta}}{2}}.
\end{equation}
\end{theorem}
\begin{proof}
The result follows directly from Theorem~\ref{theorem:main} and
Lemmas~\ref{lemma:diff(phi)} and \ref{lemma:expect(phi)}.
\end{proof}
The bound of Theorem~\ref{thm:generalization} is a general bound that
applies to {\em any} transductive algorithm. To apply it, the
stability coefficient $\beta$, which depends on $m$ and $u$, needs to
be determined. In the subsequent sections, we derive bounds on $\beta$
for a number of transductive regression algorithms
\citep{Cortes&Mohri2006,belkin04,WuS07,ZhouBLWS03,ZhuGL03}. Note that
when $\beta = O(1/\min(m, u))$, the slack term of this bound is in
$O(1/\sqrt{\min(m, u)})$.

\section{Stability of local transductive regression algorithms}
\label{sec:stability of local transductive regression algorithms}

This section describes and analyzes a general family of local
transductive regression algorithms (\LTR) generalizing the algorithm of
\cite{Cortes&Mohri2006}.

\subsection{Local transductive regression algorithms}

\LTR\ algorithms can be viewed as a generalization of the so-called
kernel regularization-based learning algorithms to the transductive
setting. The objective function that is minimized is of the form:
\begin{equation}
\label{eqn:obj}
F(f, S) = \normK{f}^2 + \frac{C}{m} \sum_{k=1}^m
c(f, x_k) + \frac{C'}{u} \sum_{k=1}^u \widetilde{c}(f, x_{m+k}),
\end{equation}
where $\normK{\cdot}$ is the norm in the reproducing kernel Hilbert
space (RKHS) with associated kernel $K$, $C \geq 0$ and $C' \geq 0$
are trade-off parameters, $f$ is the hypothesis and $\widetilde{c} (f,
x) = (f(x) - \widetilde{y}(x))^2$ is the error of $f$ on the unlabeled
point $x$ with respect to a pseudo-target $\widetilde{y}$.

Pseudo-targets are obtained from neighborhood labels $y(x)$ by a local
weighted average or other regression algorithms applied
locally. Neighborhoods can be defined as a ball of radius $r$ around
each point in the feature space. We will denote by $\bloc$ the
score-stability coefficient
(Definition~\ref{defn:score-beta-stability}).

\subsection{Generalization bounds}

In this section, we use the bounded-labels assumption, that is we
shall assume that for all $x\in S$, $|y(x)|\leq M$ for some $M \!>\!
0$.  \ignore{THIS IS NEVER USED Then, for any $x\in S$ the
  pseudo-target $\tilde{y}(x)$ assigned by the local estimator to an
  unlabeled example satisfies $|\tilde{y}(x)|\leq M$. This assumption
  is quite mild and is satisfied when the pseudo-targets are computed
  by Nadaraya-Watson estimators or by methods such as kernel ridge
  regression.  } We also assume that for any $x \in X$, $K(x, x) \leq
\kappa^2$.  We will use the following bound based on the reproducing
property and the Cauchy-Schwarz inequality valid for any hypothesis $h
\in H$, and for all $x \in X$,
\begin{equation}
\label{eq:kernel-bounded}
| h(x) | = |\langle h , K(x, \cdot) \rangle| \leq
\normK{h} \sqrt{K(x, x)} \leq \kappa \normK{h}.
\end{equation}

\begin{lemma}
\label{lem:max_value} 
Let $h$ be the hypothesis minimizing (\ref{eqn:obj}). Assume that
for any $x\in X$, $K(x,x) \leq \kappa^2$. Then, for any $x \!\in\! X$,
$|h(x)| \!\leq\! \kappa M\sqrt{C + C'}$.
\end{lemma}
\begin{proof}
  The proof is an adaptation of the technique of~\cite{bousquet-jmlr}
  to \LTR\ algorithms. By Equation~\ref{eq:kernel-bounded}, $|h(x)|
  \leq \kappa \normK{h}$. Let $\mat{0} \in \mathbb{R}^{m+u}$ be the
  hypothesis assigning label zero to all examples. By the definition
  of $h$,
\begin{equation} 
F(h,S) \leq F(\mat{0},S) \leq (C+C')M^2.
\end{equation}
Using the fact that $\normK{h}\leq \sqrt{F(h,S)}$
yields the statement of the lemma.
\end{proof}
Since $|h(x)| \leq \kappa M\sqrt{C+C'}$, this immediately gives us a
bound on $|h(x) - y(x)|$:
\begin{equation}
\label{eq:25}
\abs{h(x) - y(x)} \leq M(1 + \kappa\sqrt{C + C'}),
\end{equation}
and we are in a position to apply Theorem~\ref{thm:generalization}
with $B = AM$, $A = 1 + \kappa\sqrt{C + C'}$.

Let $h$ be a hypothesis obtained by training on $S$ and $h'$ by
training on $S'$. To determine the cost-stability coefficient $\beta$,
we must upper-bound $|c(h, x) - c(h', x)|$. Let $\Delta h = h -
h'$. Then, for all $x \in X$,
\begin{align}
  |c(h, x) - c(h', x)| & = \Big|\Delta h (x) \left[ (h(x) - y(x)) +
    (h'(x) - y(x)) \right]\Big| \\
  & \leq 2M (1 + \kappa \sqrt{C + C'}) |
  \Delta h (x) |.
\end{align}
As in Inequality~\ref{eq:kernel-bounded}, for all $x \in X$, $|\Delta
h(x)| \leq \kappa \normK{\Delta h}$, thus for all $x \in X$,
\begin{equation}
\label{eq:delta c}
|c(h, x) - c(h', x)| \leq 2 M (1 + \kappa \sqrt{C + C'})
\kappa \normK{\Delta h}.  
\end{equation}
It remains to bound $\normK{\Delta h}$. Our approach towards bounding
$\normK{\Delta h}$ is similar to the one used by \cite{bousquet00},
and relies on the convexity of $h \mapsto c(h, x)$. Note however, that
in the case of $\widetilde c$, the pseudo-targets may depend on the
training set $S$. This dependency matters when we wish to apply
convexity of two hypotheses $h$ and $h'$ obtained by training on
different samples $S$ and $S'$.  For convenience, for any two such
fixed hypotheses $h$ and $h'$, we extend the definition of $\widetilde
c$ as follows. For all $t \in [0, 1]$,
\begin{equation}
\widetilde{c} (th + (1 - t)h', x) = \big( (th + (1 - t) h') (x) - (t
\widetilde{y} + (1 - t) \widetilde{y}')\big)^2.
\end{equation}
This allows us to use the same convexity property for $\widetilde c$
as for $c$ for any two fixed hypotheses $h$ and $h'$ as verified by
the following lemma.

\begin{lemma}
\label{lemma:c-tilde-convex}
Let $h$ be a hypothesis obtained by training on $S$ and $h'$ by
training on $S'$. Then, for all $t \in [0, 1]$,
\begin{equation}
t \widetilde c (h, x) + (1 - t) \widetilde c(h', x) \geq \widetilde c
(th + (1 - t)h', x).
\end{equation}
\end{lemma}
\begin{proof}
Let $\widetilde{y} = \widetilde y(x)$ be the pseudo-target value at
$x$ when the training set is $S$ and $\widetilde{y}' = \widetilde
y'(x)$ when the training set is $S'$. For all $t \in [0, 1]$,
\begin{align*}
& t c (h, x) + (1 - t) c(h', x) - c (th + (1 - t)h', x) \\
= \ & t (h(x) - \widetilde{y})^2 + (1 - t) (h'(x) -
\widetilde{y}')^2 - \left[(th(x) + (1 - t)h'(x) - (t\widetilde{y}
+ (1 - t)\widetilde{y}')\right]^2 \\
= \ & t (h(x) - \widetilde{y})^2 + (1 - t) (h'(x) -
\widetilde{y}')^2 - \left[t(h(x) - \widetilde{y}) + (1 -
t)(h'(x) - \widetilde{y}')\right]^2.
\end{align*}
The statement of the lemma follows directly by the convexity of the
function $x \mapsto x^2$ defined over $\Rset$. 
\end{proof}
Recall that $\bloc$ denotes the score-stability of the algorithm that
produces the pseudo-targets. In Lemma~\ref{lemma:norm-lemma} we
present an upper-bound $\normK{\Delta h}$, which can then be plugged
into Equation~\ref{eq:delta c} to determine the stability of \LTR.
\begin{lemma}
\label{lemma:bounded_labels} 
Assume that for all $x \in X$, $|y(x)| \leq M$. Let $S$ and $S'$ be
two samples differing by exactly one point. Let $h$ be the hypothesis
returned by the algorithm minimizing the objective function $F(f, S)$,
$h'$ be the hypothesis obtained by minimization of $F(f, S')$ and let
$\widetilde y$ and $\widetilde y'$ be the corresponding
pseudo-targets. Then for all $i \in [1, m + u]$, 
\begin{align}
\label{eqn:lemma1.5}
&
\frac{C}{m} \left[ c(h', x_i) - c(h, x_i) \right] + \frac{C'}{u}
\left[ \widetilde{c}(h', x_i) - \widetilde{c}(h, x_i) \right]
\nonumber \\
& \leq
2AM\left( \kappa \normK{\Delta h} \left( \frac{C}{m} +
    \frac{C'}{u}\right) +\bloc \frac{C'}{u}\right),
\end{align}
where $\Delta h = h' - h$ and $A = 1+\kappa \sqrt{C+C'}$.
\end{lemma}
\begin{proof} 
  From Equation~\ref{eq:delta c}, we know that: 
  \begin{equation}|c(h', x_i) - c(h, x_i)| \leq 2 M (1 + \kappa \sqrt{C + C'})
  \kappa \normK{\Delta h}. \end{equation} It remains to bound $|\widetilde{c}(h',
  x_i) - \widetilde{c}(h, x_i)|$.
  \begin{eqnarray*}
    \widetilde{c}(h', x_i) - \widetilde{c}(h, x_i) & = & (h'(x) -
    \widetilde{y}'(x))^2 - (h(x) - \widetilde{y}(x))^2 \\ & = & \left( 
      (h'(x) - \widetilde{y}'(x)) + (h(x) - \widetilde{y}(x)) \right) \left(
      \Delta h (x) - (\widetilde{y}'(x) - \widetilde{y}(x)) \right) \\ &
    \leq & 2M (1 + \kappa \sqrt{C + C'}) \left( \kappa \normK{\Delta h} +
      \bloc \right)
  \end{eqnarray*}
  Here, we are using score-stability $\bloc$ of the local algorithm in
  $|\widetilde{y}'(x) - \widetilde{y}(x)| \leq \bloc$ and that $|h(x)
  - \widetilde{y}(x)| \leq M(1 + \kappa\sqrt{C + C'})$ when
  $|\widetilde{y}(x)| \leq M$ (by Lemma~\ref{lem:max_value}).

  Plugging the bounds for $|c(h', x_i) - c(h, x_i)|$ and
  $|\widetilde{c}(h', x_i) - \widetilde{c}(h, x_i)|$ into the left
  hand side of Equation~\ref{eqn:lemma1.5} yields the statement of the
  lemma.
\end{proof}
\begin{lemma}
\label{lemma:norm-lemma}
Assume that for all $x \in X$, $|y(x)| \leq M$. Let $S$ and $S'$ be
two samples differing by exactly one point. Let $h$ be the hypothesis
returned by the algorithm minimizing the objective function $F(f, S)$,
$h'$ the hypothesis obtained by minimization of $F(f, S')$ and let
$\widetilde y$ and $\widetilde y'$ be the corresponding
pseudo-targets. Then 
\begin{equation}
\normK{\Delta h}^2 \leq  2AM\left( \kappa \normK{\Delta h} \left( \frac{C}{m} +
    \frac{C'}{u}\right) +\bloc \frac{C'}{u}\right),
\end{equation}
where $\Delta h = h' - h$ and $A = 1+\kappa \sqrt{C+C'}$.
\end{lemma}
\begin{proof}
By the definition of $h$ and $h'$, we have
\begin{equation*}
h = \argmin_{f \in H} F(f, S) \quad \mbox{and} \quad h' = \argmin_{f
\in H} F (f, S').
\end{equation*} 
Let $t \in [0, 1]$. Then $h + t \Delta h$ and $h' - t \Delta h$
satisfy:
\begin{eqnarray}
\label{eqn:h_t_delta}
F(h, S) - F(h + t \Delta h, S) & \leq & 0 \\
\label{eqn:h'_t_delta}
F(h', S') - F(h' - t \Delta h, S') & \leq & 0
\end{eqnarray} 
For notational ease, let $h_{t\Delta}$ denote $h + t \Delta h$ and
$h'_{t\Delta}$ denote $h' - t \Delta h$. Summing the two
inequalities in Equations~\ref{eqn:h_t_delta} and \ref{eqn:h'_t_delta}
yields:
\begin{align*}
  & \frac{C}{m}\sum_{k=1}^m \left[c(h, x_k) - c(h_{t\Delta}, x_k)
  \right] + \frac{C'}{u} \sum_{k=1}^u \left[ \widetilde{c}(h,
    x_{m+k}) - \widetilde{c}(h_{t\Delta}, x_{m+k}) \right] + \\
  & \frac{C}{m} \sum_{k=1, k \neq i}^m \left[ c (h', x_k) -
    c(h'_{t\Delta}, x_{k}) \right] + \frac{C'}{u} \sum_{k=1, k \neq
    j}^u \left[ \widetilde{c} (h', x_{m+k}) - \widetilde{c}
    (h'_{t\Delta}, x_{m+k}) \right] + \\
  & \frac{C}{m} \left[ c (h', x_{m+j}) - c(h'_{t\Delta}, x_{m+j})
  \right] + \frac{C'}{u} \left[ \widetilde{c} (h', x_{i}) -
    \widetilde{c} (h'_{t\Delta}, x_{i}) \right] +
  \\
& \normK{h}^2 - \normK{h_{t\Delta}}^2 +
\normK{h'}^2 - \normK{h'_{t\Delta}}^2 \quad \leq 0.
\end{align*}
By the convexity of $c(h, \cdot)$ in $h$, it follows that for all $k
\in [1, m + u]$
\begin{equation}
c(h, x_k) - c(h_{t\Delta}, x_k) \geq t \left[c(h, x_k) - c(h +
\Delta h, x_k)\right],
\end{equation}  
and
\begin{equation}
c(h', x_k) - c(h'_{t\Delta}, x_k) \geq t \left[c(h',
x_k) - c(h' - \Delta h, x_k)\right].
\end{equation} 
By Lemma~\ref{lemma:c-tilde-convex}, similar inequalities hold for
$\widetilde{c}$. These observations lead to:
\begin{align*}
& \frac{Ct}{m}\sum_{k=1}^m
\left[c(h, x_k) - c(h', x_k) \right] + \frac{C't}{u}
\sum_{k=1}^u \left[ \widetilde{c}(h, x_{m+k}) -
\widetilde{c}(h', x_{m+k}) \right] + \\
& \frac{Ct}{m} \sum_{k=1, k \neq i}^m \left[ c
(h', x_k) - c(h, x_{k}) \right] + \frac{C't}{u} \sum_{k=1, k
\neq j}^u \left[ \widetilde{c}(h', x_{m+k}) - \widetilde{c}(h,
x_{m+k}) \right] + \\ 
& \frac{Ct}{m} \left[ c (h', x_{m+j}) - c(h,
x_{m+j}) \right] + \frac{C't}{u} \left[ \widetilde{c}(h',
x_{i}) - \widetilde{c}(h, x_{i}) \right] + \\
& \normK{h}^2 - \normK{h_{t\Delta}}^2 +
\normK{h'}^2 - \normK{h'_{t\Delta}}^2 \quad \leq 0.
\end{align*}
Let $E$ denote $\normK{h}^2 - \normK{h_{t\Delta}}^2 + \normK{h'}^2
- \normK{h'_{t\Delta}}^2$. Simplifying the previous inequality
leads to:
\begin{eqnarray*}
E & \leq & \frac{Ct}{m} \left[ c(h', x_i) - c(h, x_i) + c(h, x_{m+j})
- c(h', x_{m+j}) \right] - \\
& & \frac{C't}{u} \left[ \widetilde{c}(h', x_i) - \widetilde{c}(h,
x_i) + \widetilde{c}(h, x_{m+j}) - \widetilde{c}(h', x_{m+j}) \right].
\end{eqnarray*}
Let $A = 1+\kappa \sqrt{C+C'}$. Using Lemma~\ref{lemma:bounded_labels}
twice (with $x_i$ and $x_{m+j}$), the expression above can be bounded
by
\begin{eqnarray}
\label{eqn:bounded-e}
E & \leq & 4AMt\left( \kappa \normK{\Delta h} \left( \frac{C}{m} +
    \frac{C'}{u}\right) +\bloc \frac{C'}{u}\right).
\end{eqnarray}
Finally, since $\|h\|_K^2 = \langle
h, h \rangle_K$ for any $h \in H$, it is not hard to show that:
\begin{equation}
\label{eqn:proof-2}
\normK{h}^2 - \normK{h + t \Delta h}^2 + \normK{h'}^2 -
\normK{h' - t \Delta h}^2 = 2t \normK{\Delta h}^2 (1 - t).
\end{equation}
Using Equation~\ref{eqn:proof-2} in Equation~\ref{eqn:bounded-e}, it
follows that: 
\begin{equation} \normK{\Delta h}^2 (1 - t) \leq 2AM\left( \kappa \normK{\Delta h}
  \left( \frac{C}{m} + \frac{C'}{u}\right) +\bloc
  \frac{C'}{u}\right). \end{equation} 
Taking the limit as $t \to 0$ yields the statement of the lemma.
\end{proof}

The following is the main result of this section, a stability-based
generalization bound for \LTR.
\begin{theorem}
\label{thm:treg_stab}
Assume that for all $x \in X$, $|y(x)| \leq M$ and there exists
$\kappa$ such that for all $x \in X$, $K(x, x) \leq \kappa^2$.
Further, assume that the local estimator has score-stability
$\bloc$. Let $A= 1+ \kappa \sqrt{C+C'}$.  Then, \LTR\ is uniformly
$\beta$-cost-stable with
\begin{equation*}
  \beta \leq {2(AM)^2 \kappa^2} \biggl[ \frac{C}{m} + \frac{C'}{u} +
  \sqrt{\biggl(\frac{C}{m} + \frac{C'}{u}\biggr)^2 + \frac{2C'
      \bloc}{AM \kappa^2 u}}\biggr]. 
\end{equation*}
\end{theorem}
\begin{proof}
  From Lemma~\ref{lemma:norm-lemma}, we know that 
  \begin{equation} \normK{\Delta h}^2 \leq 2AM\left( \kappa \normK{\Delta h} \left(
      \frac{C}{m} + \frac{C'}{u}\right) +\bloc \frac{C'}{u}\right), \end{equation}
  where $\Delta h = h' - h$ and $A = 1+\kappa \sqrt{C+C'}$.  This
  implies that $\norm{\Delta h}$ is bounded by the non-negative root
  of the second-degree polynomial which gives
  \begin{equation} \normK{\Delta h} \leq AM \kappa \Biggr[ \left(\frac{C}{m} +
    \frac{C'}{u}\right) + \sqrt{\left(\frac{C}{m} +
      \frac{C'}{u}\right)^2 + \frac{2C'\bloc}{A M \kappa^2 u}}
  \Biggr]. \end{equation} Using the above bound on $\normK{\Delta h}$ in
  Equation~\ref{eq:delta c} yields the desired bound on the stability
  coefficient of \LTR\ and completes the proof. 
\end{proof}
Our experiments with local transductive regression in
Section~\ref{sec:experiments} will show the benefit of this bound
for model selection.

\section{Stability of unconstrained regularization algorithms}
\label{sec:unconstrained}

\subsection{Unconstrained regularization algorithms}

In this section, we consider a family of transductive regression
algorithms that can be formulated as the following optimization
problem:
\begin{equation}
\label{eqn:obj_fun}
\min_{\mh} \mh^\top \mat{Q} \mh + (\mh - \y)^\top
\mat{C} (\mh - \y),
\end{equation}
where $\mat{Q} \in \Rset^{(m + u) \times (m + u)}$ is a symmetric
regularization matrix, $\mat{C} \in \Rset^{(m + u) \times (m + u)}$ a
symmetric matrix of empirical weights (in practice it is often a
diagonal matrix), $\y \in \Rset^{(m + u) \times 1}$ the target
values of the $m$ labeled points together with the pseudo-target
values of the $u$ unlabeled points (in some formulations, the
pseudo-target value is 0), and $\mh \in \Rset^{(m + u) \times 1}$
a column matrix whose $i$th row is the predicted target value for the
$x_i$. The closed-form solution of (\ref{eqn:obj_fun}) is given by
\begin{equation}
\label{eqn:closed_form_main}
\mh = (\mat{C}^{-1} \mat{Q} + \I)^{-1} \y.
\end{equation}
The formulation (\ref{eqn:obj_fun}) is quite general and includes as
special cases the algorithms of
\cite{belkin04,WuS07,ZhouBLWS03,ZhuGL03}. We present a general
framework for bounding the stability coefficient of these algorithms
and then examine the stability coefficient of each of these algorithms
in turn. \ignore{\footnote{If we allow matrix $\mat{Q}$ to depend on
    the training/test partition, then our formulation also covers the
    algorithm of \cite{Joachims03}, which is based on spectral graph
    partitioning.  However, we were not able to obtain fully explicit
    bounds on the stability coefficient for this algorithm within our
    framework.}}

\subsection{Score-based stability analysis}

For a symmetric matrix $\mat{A} \in \Rset^{n\times n}$ we denote by
$\lambda_M(\mat{A})$ its largest and by $\lambda_m (\mat{A})$ its
smallest eigenvalue. Thus, for any $\mat{v} \in \Rset^{n \times 1}$,
$\lambda_m(\mat{A}) \ltwo{\mat{v}} \leq \|\mat{A}\mat{v}\|_2\leq
\lambda_M(\mat{A}) \|\mat{v}\|_2$. We will also use, in the proof of
the following proposition, the fact that for symmetric matrices
$\mat{A}, \mat{B} \in \Rset^{n \times n}$,
$\lambda_M(\mat{A}\mat{B})\leq \lambda_M(\mat{A})\lambda_M(\mat{B})$.

\begin{theorem}
  \label{prop:closed_form}
  Let $\mh^*$ and $\mh'^*$ solve (\ref{eqn:obj_fun}), under
  test and training sets that differ exactly in one point and let
  $\mat{C}, \mat{C}', \y, \y'$ be the corresponding
  empirical weight and the target value matrices. Then,
  \begin{equation}
    \label{eqn:delta_h_bound}
    \normtr{\mh^* - \mh'^*}_\infty \leq \normtr{\mh^* -
      \mh'^*}_2 \leq \frac{\normtr{\y -
        \y'}_2}{\frac{\lambda_m (\mat{Q})}{\lambda_M (\mat{C})} + 1} +
    \frac{ \lambda_M(\mat{Q}) \normtr{\mat{C}'^{-1} - \mat{C}^{-1}}_2 \, \normtr{\y'}_2}{
      \Big(\frac{\lambda_m (\mat{Q})}{\lambda_M (\mat{C}')} + 1 \Big)
      \Big(\frac{\lambda_m (\mat{Q})}{\lambda_M (\mat{C})} + 1 \Big)}
  \end{equation}
\end{theorem}
\begin{proof}
  The first inequality holds as a result of the general relation
  between norm-infinity and norm-2.  Let $\Delta \mh^*= \mh^*
  - \mh'^*$ and $\Delta \y = \y - \y'$. By
  definition,
\begin{align}
  \Delta \mh^* = & (\mat{C}^{-1} \mat{Q} + \I)^{-1} \y -
  (\mat{C}'^{-1} \mat{Q} + \I)^{-1} \y' \\
  = & (\mat{C}^{-1} \mat{Q} + \I)^{-1} \Delta \y +
  ((\mat{C}^{-1} \mat{Q} + \I)^{-1} - (\mat{C}'^{-1} \mat{Q} +
  \I)^{-1}) \y' \\
  = & (\mat{C}^{-1} \mat{Q} + \I)^{-1} \Delta \y 
   + \left[(\mat{C}'^{-1} \mat{Q} + \I)^{-1} \left[(\mat{C}'^{-1} -
    \mat{C}^{-1}) \mat{Q}\right] (\mat{C}^{-1} \mat{Q} +
  \I)^{-1}\right] \y'. 
  \label{eqn:hpmh}
\end{align}
Since $\ltwo{[\mat{C}^{-1} \mat{Q} + \I]^{-1} } =
\lambda_{\max}([\mat{C}^{-1} \mat{Q} + \I]^{-1}) =
\lambda_{\min}(\mat{C}^{-1} \mat{Q} + \I)$, and $\lambda_m
(\mat{C}^{-1} \mat{Q} + \I) \geq \frac{\lambda_m
  (\mat{Q})}{\lambda_M (\mat{C})} + 1$, $\ltwo{\Delta \mh^*}$ can be
bounded as follows:
  \begin{align}
    \label{eqn:bound_on_delta_h}
    \ltwo{\mat{\Delta h^*}} 
& \leq \frac{\ltwo{\Delta \y}}{\lambda_m
      (\mat{C}^{-1} \mat{Q} + \I)} + \frac{ \lambda_M(\mat{Q}) \ltwo{\mat{C}'^{-1} -  \mat{C}^{-1}} \, \ltwo{\y'}}{\lambda_m (\mat{C}'^{-1}
      \mat{Q} + \I) \lambda_m (\mat{C}^{-1} \mat{Q} + \I)}\\
& \leq \frac{\ltwo{\Delta \y}}{\frac{\lambda_m (\mat{Q})}{\lambda_M (\mat{C})} + 1} +
    \frac{ \lambda_M(\mat{Q}) \ltwo{\mat{C}'^{-1} - \mat{C}^{-1}} \, \ltwo{\y'}}{
      \Big(\frac{\lambda_m (\mat{Q})}{\lambda_M (\mat{C}')} + 1 \Big)
      \Big(\frac{\lambda_m (\mat{Q})}{\lambda_M (\mat{C})} + 1 \Big)}. 
  \end{align}
This proves the second inequality.
\end{proof}
The theorem helps derive score-stability bounds for various
transductive regression algorithms \citep{ZhouBLWS03,WuS07,ZhuGL03}
based on the closed-form solution for the hypothesis. Recall that
score-stability (Definition~\ref{defn:score-beta-stability}) is the
maximum change in the hypothesis score on any point $x$ as the
learning algorithm is trained on two training sets that differ in
exactly one point, that is precisely an upper-bound on
$\normtr{\mh^* - \mh'^*}_\infty$.

\subsection{Application}

For each of the algorithms in \citep{ZhouBLWS03,WuS07,ZhuGL03}, an
estimate of 0 is used for unlabeled points. Thus, the vector $\y$
has the following structure: the entries corresponding to training
examples are their true labels and those corresponding to the
unlabeled examples are $0$. 

For each one of the three algorithms, we make the bounded labels
assumption (for all $x \in X, |y(x)| \leq M$ for some $M > 0$).  It is
then not difficult to show that $\ltwo{\y - \y'} \leq
\sqrt{2}{M}$ and $\ltwo{\y'} \leq \sqrt{m} M$. Furthermore, all
the stability bounds derived are based on the notion of
score-stability (Definition~\ref{defn:score-beta-stability}).

\subsubsection{Consistency method ($\CM$)}
In the $\CM$ algorithm~\citep{ZhouBLWS03}, the matrix $\mat{Q}$ is a
normalized Laplacian of a weight matrix $\mat{W}\in \Rset^{(m + u)
  \times (m + u)}$ that captures affinity between pairs of points in
the full sample $X$. Thus,
$\mat{Q}=\I-\mat{D}^{-1/2}\mat{W}\mat{D}^{-1/2}$, where $\mat{D}
\in \Rset^{(m + u) \times (m + u)}$ is a diagonal matrix, with
$[\mat{D}]_{i,i} = \sum_{j} [\mat{W}]_{i,j}$.  Note that $\lambda_m
(\mat{Q}) = 0$. Furthermore, matrices $\mat{C}$ and $\mat{C}'$ are
identical in $\CM$, both diagonal matrices with $(i, i)$th entry equal
to a positive constant $\mu > 0$. Thus $\mat{C}^{-1} = \mat{C}'^{-1}$
and using Proposition~\ref{prop:closed_form}, we obtain the following bound
on the score-stability of the $\CM$ algorithm: $\beta_{\CM} \leq
\sqrt{2} M$.

\subsubsection{Local learning regularization ($\LLR$)}
\label{subsec:llr}
In the $\LLR$ algorithm ~\citep{WuS07}, the regularization matrix
$\mat{Q}$ is $(\I-\mat{A})^\top(\I-\mat{A})$, where $\I
\in \Rset^{(m + u) \times (m + u)}$ is an identity matrix and $\mat{A}
\in \Rset^{(m + u) \times (m + u)}$ is a non-negative weight matrix
that captures the local similarity between all pairs of points in
$X$. $\mat{A}$ is normalized, i.e.~each of its rows sum to $1$.  Let
$C_l, C_u>0$ be two positive constants. The matrix $\mat{C}$ is a
diagonal matrix with $[\mat{C}]_{i,i} = C_l$ if $x_i \in S$ and $C_u$
otherwise.  Let $C_{\max}=\max\{C_l,C_u\}$ and
$C_{\min}=\min\{C_l,C_u\}$. Thus, $ \norm{\mat{C}'^{-1} -
  \mat{C}^{-1}}_2 = \sqrt{2} \left(\frac{1}{C_{\min}} -
  \frac{1}{C_{\max}} \right)$. By the Perron-Frobenius theorem, its
eigenvalues lie in the interval $(-1,1]$ and $\lambda_M(\mat{A}) \leq
1$. Thus, $\lambda_m (\mat{Q}) \geq 0$ and $\lambda_M (\mat{Q}) \leq
4$ and we have the following bound on the score-stability of the
$\LLR$ algorithm: $\beta_{\LLR} \leq \sqrt{2} M + 4\sqrt{2m} M\left(
  \frac{1}{C_{\min}} - \frac{1}{C_{\max}} \right) \leq \sqrt{2}M +
\frac{4\sqrt{2m} M}{C_{\min}}$.

\subsubsection{Gaussian Mean Fields algorithm} $\GMF$ \citep{ZhuGL03} is
  very similar to the $\LLR$, and admits exactly the same
  stability coefficient.

  Thus, using our bounding technique, the stability coefficients of
  the algorithms of $\CM, \LLR$, and $\GMF$ can be large. Without
  additional constraints on the matrix $\mat{Q}$, these algorithms do
  not seem to be stable enough for the generalization bound of
  Theorem~\ref{thm:generalization} to converge. The next section in
  fact demonstrates that by presenting a constant lower bound on their
  score-stability.

\subsection{Lower bound on stability coefficient}
\label{sec:lower_bound}

The stability coefficient is a function of the sample size. For
stability learning bounds to converge, it must go to zero as a
function of the sample size. The following theorem proves that the
stability coefficient of the $\CM$ algorithm is lower-bounded by a
constant for some problems. A similar lower bound can be given for
the other two algorithms examined.

\begin{theorem}
  There exists a transductive regression problem with $m \!\geq\! 2$
  labeled samples and $m$ unlabeled samples and a diagonal matrix
  $\mat{C}$ for which the score-stability $\beta$ of the $\CM$
  algorithm admits the following lower bound:
\begin{equation}
\beta \geq \frac{1}{2} \frac{C}{C + 1}.
\end{equation}
\end{theorem}

\begin{proof}
  Consider a transductive regression problem with $2m$ instances where
  $m$ instances have a target value of 0 and the other $m$ instances a
  target value of 1. Let the labeled sample $S$ include exactly the
  instances $x_1, \ldots, x_m$ with target value 0 and $U$ be defined
  by the complement $x_{m + 1}, \ldots, x_{2m}$. Let $\L$ denote an $m
  \times m$ normalized graph Laplacian matrix, with 1s along the
  diagonal and all off-diagonal terms equal to $-\frac{1}{m -
    1}$. Then the matrix $\mat{Q}$ is defined with the following block
  structure:
  \begin{equation} \mat{Q} = 
    \begin{bmatrix}
      \L & \mat{0} \\ \noalign{\medskip}
      \mat{0} & \L 
    \end{bmatrix}
  \end{equation}
  In our example, we set $\mat{C}$ to be a diagonal matrix with all
  its entries equal to the constant $C$. The matrix $\M =
  \mat{C}^{-1} \mat{Q} + \I$ has the following block structure:
  \begin{equation} 
    \M =
    \begin{bmatrix}
      \mat{N} & \mat{0} \\ \noalign{\medskip}
      \mat{0} & \mat{N} 
    \end{bmatrix},
  \end{equation}
  where $\mat{N}$ is the $m \times m$ matrix whose diagonal entries
  are all equal to  $1 + \frac{1}{C}$ and whose off-diagonal entries
  all equal to $-\frac{1}{C(m - 1)}$.

  Now, consider the training sample $S'$ obtained from $S$ by swapping
  a labeled point with an unlabeled point. For the sake of
  convenience, let the index of this point be $m$. The $\y$
  vector changes (to $\mat{y'}$) only in the $m$th position. Thus, all
  the entries of $\Delta \y = \y - \mat{y'}$ are zero except
  from its $m$th entry which equals 1.  By Equation~\ref{eqn:hpmh},
  $\Delta \mh^* = \M^{-1} \Delta \y$, thus, $\Delta
  \mh^*$ is exactly the $m$th column of $\M^{-1}$. Let
  $[\M^{-1}]_{m,m} = [\mat{N}^{-1}]_{m,m}$ denote the $(m, m)$
  entry of $\M^{-1}$ which coincides with the the $(m, m)$ entry
  of $\mat{N}^{-1}$. Since $\ltwo{\Delta \mh^*} \geq
  |[\M^{-1}]_{m,m}|$, to give a lower bound on $\ltwo{\Delta
    \mh^*}$, it suffices to lower bound $|[\mat{N}^{-1}]_{m,m}|$.
  To do so, we can compute $[\mat{N}^{-1}]_{m,m}$.
  
  By symmetry, the diagonal entries of $\mat{N}^{-1}$ are all equal to
  some value $a$, thus $ma = \Tr(\mat{N}^{-1})$, which can be computed
  from the inverses of the eigenvalues of $\mat{N}$. Observe that
  $\mat{N}_0 = \mat{N} - (1 + \frac{1}{C} + \frac{1}{C(m - 1)})
  \I$ is a matrix with all entries equal to $- \frac{1}{C(m -
    1)}$.  Thus, it is a rank one matrix and its only non-zero
  eigenvalue coincides with its trace: $\Tr(\mat{N}_0) = - \frac{m}{C(m
    - 1)}$. Since $1 + \frac{1}{C} + \frac{1}{C(m - 1)} = \frac{(m -
    1) C + m}{(m - 1)C}$, this shows that the eigenvalues of $\mat{N}$
  are $\frac{m}{C(m - 1)} + \frac{(m - 1) C + m}{(m - 1)C} = 1$ with
  multiplicity $1$, and $\frac{(m - 1) C + m}{(m - 1)C}$ with
  multiplicity $m - 1$. Thus, $ma = \Tr(\mat{N}^{-1}) = 1 + \frac{(m -
    1)^2 C}{(m - 1)C + m}$, which gives
\begin{equation}
  a = \frac{1}{m} + \frac{\frac{m - 1}{m} C}{C + \frac{m}{m - 1}} \geq
  \frac{\frac{1}{2} C}{C + 1},
\end{equation}
since for $m \geq 2$, $\frac{1}{2} \leq \frac{m - 1}{m} \leq 1$.
\end{proof}
An example of constraint that can help guarantee stability is the
condition $\sum_{i=1}^{m+u} h (x_i) = 0$ used in the algorithm of
\cite{belkin04}. In the next section, we give a generalization bound
for a family of algorithms based on this constraint this and then
describe a general method for making the algorithms just examined
stable.

\section{Stability of constrained regularization algorithms}
\label{sec:constrained}

\subsection{Constrained graph regularization algorithms}
\label{sec:constrained_algos}

Here, we examine constrained regularization algorithms such as the
graph Laplacian regularization algorithm of \cite{belkin04}. Given a
weighted graph $G \!=\! (X, E)$ in which edge weights can be
interpreted as similarities between vertices, the task consists of
predicting the vertex labels. The input space $X$ is thus reduced to
the set of vertices, and a hypothesis $h\colon X \to \Rset$ can be
identified with the finite-dimensional vector $\mh$ of its
predictions $\mh \!=\! [h(x_1), \ldots, h(x_{m + u})]^\top$. The
hypothesis set $H$ can thus be identified with $\Rset^{m + u}$
here. Let $\mh_S$ denote the restriction of $\mh$ to the
training points, $[h(x_1), \ldots, h(x_{m})]^\top \!\in\!  \Rset^m$,
and similarly let $\y_S$ denote $[y_1, \ldots, y_m]^\top \!\in\!
\Rset^m$.

The general family of constrained graph regularization algorithms can
then be defined by the following optimization problem:
\begin{align}
\label{eqn:belkin}
\min_{\mh \in H} & \ \mh^\top \L \mh + \frac{C}{m} (\mh_S - \y_S)^\top (\mh_S - \y_S)\\
\text{subject to:} & \ \mh^\top \u = 0, \nonumber
\end{align}
where $\L \!\in\! \Rset^{(m + u) \times (m + u)}$ is a positive
semi-definite symmetric matrix, $y_i$, $i \in [1, m]$, the target
values of the $m$ labeled nodes, and $\u \in \Rset^{m + u}$ a fixed
vector. The constraint of the optimization thus restricts the space of
solutions to be in $H_1$, the hyperplane in $H$ of the vectors
orthogonal to $\u$. We denote by $\P$ the projection matrix over the
hyperplane $H_1$. As further discussed later, for stability reasons,
$\u$ is typically selected to be orthogonal to the range of $\L$,
$\range(\L)$. More generally, the optimizations constraint can be
generalized to orthogonality with respect to a subspace $U$ such that
the space of solutions $H_1$ be a subset of $\range(\L)$.

In the case of the regularization algorithm of \cite{belkin04}, $\L$
is the graph Laplacian. Thus, $\mh^\top \L \mh = \sum_{ij = 1}^m
w_{ij} (h(x_i) - h(x_j))^2$, for some weight matrix $(w_{ij})$. The
vector $\u$ is defined to be $\mat{1}$, that is all its entries equal
1. For this algorithm, the authors further assume the label vector
$\y$ to be centered, which implies that $\u^\top \y = 0$, and also
that the graph $G$ is connected. This last assumption implies that the
zero eigenvalue of the Laplacian has multiplicity one and that 
$H_1$ coincides with $\range(\L)$.

For a sample $S$ drawn without replacement from $X$, define $\I_S \in
\Rset^{(m + u) \times (m + u)}$ as the diagonal matrix with
$[\I_S]_{i,i} = 1$ if $x_i \in S$ and 0 otherwise. Similarly, let
$\y_S \in \Rset^{(m + u) \times 1}$ be the vector with
$[\y_S]_{i,1} = y_i$ if $x_i \in S$ and 0 otherwise. Then, the
Lagrangian associated to the problem~(\ref{eqn:belkin}) is ${\cal L} =
\mh^\top \L \mh + \frac{C}{m} (\mh_S - \y_S)^\top
(\mh_S - \y_S) + \beta \mh^\top \u$, where $\beta \in \Rset$
is a Lagrange variable. Setting its gradient with respect to $\mh$
to zero gives
\begin{equation}
\L \mh + \frac{C}{m} (\mh_S - \y_S) + \beta \u = 0.
\end{equation}
Multiplying by the projection matrix $\P$ gives
\begin{equation}
\label{eq:57}
\P (\L + \frac{C}{m} \I_S) \mh = \frac{C}{m} \P \y_S - \beta \P \u = \frac{C}{m} \P \y_S.
\end{equation}

\subsection{Score stability of graph Laplacian regularization algorithm}
\label{sec:constrained_score}

This section gives a simple generalization bound for the graph
Laplacian regularization algorithm using a closed-form
solution of (\ref{eq:57}) and a score-stability analysis.

In the case of the graph Laplacian regularization algorithm of
\cite{belkin04} with the assumptions already indicated, matrix $\P
(\frac{m}{C} \L + \I_S)$ is invertible. Then, Equation~(\ref{eq:57})
gives the closed-form solution:
\begin{equation}
\label{eq:58}
\mh = \big[\P (\frac{m}{C} \L + \I_S)\big]^{-1} \P \y_S,
\end{equation}
which clearly verifies the constraint of the optimization problem.

\begin{theorem}
\label{thm:belkin-gen}
Assume that the graph $G \!=\! (X, E)$ is connected and that its
vertex labels are bounded: for all $x$, $|y(x)| \!\leq\! M$ for
some $M \!>\! 0$. Let $h$ denote the solution of the optimization
problem~(\ref{eqn:belkin}) where $\L$ is the graph Laplacian and $\u =
\mat{1}$, and let $A = 1 + \kappa \sqrt{C}$. Then, for any $\delta
\!>\! 0$, with probability at least $1 - \delta$,
\begin{equation}
\label{eq:bounded-belkin}
R(h) \leq  \h R(h) + \beta
 + \left(2\beta +
      \frac{(AM)^2 (m + u)}{mu} \right)
    \sqrt{\frac{\alpha (m, u) \ln {1 \over \delta}}{2}},
\end{equation}
where
\begin{equation*} 
  \alpha (m, u) = \frac{mu}{m + u - 1/2} \, \frac{1}{1 - 1 /
    (2\max\{m, u\})} \quad \text{and} \quad \beta \leq \frac{4\sqrt{2} M^2}
{m\lambda_2/C - 1} + \frac{4\sqrt{2m} M^2}{(m\lambda_2/C - 1)^2},
\end{equation*}
$\lambda_2$ being the second smallest eigenvalue of the Laplacian
$\L$.
\end{theorem}
\begin{proof}
  Our proof is similar to that of Theorem 5 in \citep{belkin04}, with
  the important exception that we no longer need to cope with vertex
  multiplicity in sampling since $S$ is sampled from $X$ without
  replacement. This makes our proof and the resulting bound
  considerably simpler and more concise.

  By Lemma~\ref{lem:max_value} and Equation~\ref{eq:25}, since the
  labels are bounded by $M$, for any $x$, the following inequality
  holds: $\abs{h(x) - y(x)} \!\leq\! M(1 + \kappa\sqrt{C}) \!=\!
  AM$. To determine the stability coefficient, it suffices to bound
  $\max_{S, S'} \linf{\mh_S - \mh_{S'}}$, where $S$ and $S'$
  are two training sets that differ only in one vertex. Let $\M_S
  = \P \left( \frac{m}{C} \L + \I_S \right)$ and
  $\M_{S'} = \P \left( \frac{m}{C} \L + \I_{S'}
  \right)$. Then,
\begin{align}
  \linf{\mh_S - \mat{h_{S'}}} & \leq \norm{\mh_S -
  \mat{h_{S'}}} \\
  & = \norm{\M_S^{-1} \P \y_S - \M_{S'}^{-1}
  \P \y_{S'}} \\
  & = \norm{\M_S^{-1} \P (\y_S - \y_{S'}) +
  (\M_S^{-1} - \M_{S'}^{-1}) \y_{S'}} \\
  & \leq \norm{\M_S^{-1} \P (\y_S - \y_{S'})} + \norm{
  (\M_S^{-1} - \M_{S'}^{-1}) \P \y_{S'}}.
\end{align}
For any column matrix $\v \in \Rset^{(m + u) \times 1}$, by the
triangle inequality and the projection property $\norm{\P \v} \leq
\norm{\v}$, the following inequalities hold:
\begin{align}
\norm{\frac{m}{C} \P \L} 
& = \norm{\frac{m}{C} \P \L + \P \I_S \v - \P \I_S \v}\\
& \leq \norm{\frac{m}{C} \P \L + \P \I_S \v}
+ \norm{\P \I_S \v}\\
& \leq \norm{\P \Big( \frac{m}{C} \L +
  \I_S \Big) \v}
+ \norm{\I_S \v}.
\end{align}
This yields the lower bound:
\begin{equation}
\norm{\M_S \v} = \norm{\P \left( \frac{m}{C} \L +
  \I_S \right) \v} \geq \frac{m}{C} \norm{\P \L} -
  \norm{\I_S \v} \geq \left( \frac{m}{C} \lambda_2 -
  1 \right) \norm{\v},
\end{equation}
which gives the following upper bound on $\norm{\M_S^{-1}},
\norm{\M_{S'}^{-1}}$:
\begin{equation}
  \norm{\M_S^{-1}} \leq \frac{1}{\frac{m}{C} \lambda_2 - 1} \quad \text{and} \quad 
  \norm{\M_{S'}^{-1}} \leq \frac{1}{\frac{m}{C} \lambda_2 - 1}.
\end{equation}
We bound each of the two terms, $\norm{\M_S^{-1} \P (\y_S - \y_{S'})}$
and $\norm{ (\M_S^{-1} - \M_{S'}^{-1}) \P \y_{S'}}$
separately. $\norm{\M_S^{-1} \P (\y_S - \y_{S'} )}$ can be bounded
straightforwardly:
\begin{equation}
  \norm{\M_S^{-1} \P (\y_S -
    \y_{S'})} \leq \norm{\M_S^{-1}} \norm{\P (\y_S -
    \y_{S'})}  \leq \norm{\M_S^{-1}} \norm{\y_S -
    \y_{S'}} \leq  \frac{\sqrt{2} M}{\frac{m}{C} \lambda_2 - 1}.
\end{equation}
$\norm{ (\M_S^{-1} - \M_{S'}^{-1}) \y_{S'}}$ is bounded as
follows:
\begin{eqnarray}
\norm{ (\M_S^{-1} - \M_{S'}^{-1})
  \P \y_{S'}} & = & \norm{ \M_{S'}^{-1} (\M_{S'} -
  \M_S) \M_S^{-1} \P \y_{S'} } \\
& = & \norm{\M_{S'}^{-1} \P (\I_{S'} - \I_S)
  \M_S^{-1} \P \y_{S'}} \\
& \leq & \frac{\sqrt{2m} M}{\left(\frac{m}{C} \lambda_2 -
  1\right)^2}.
\end{eqnarray}
This leads to the following bound on $\linf{\mh_S -
  \mh_{S'}}$:
\begin{equation}
\label{eqn:linf}
  \linf{\mh_S - \mh_{S'}} \leq \frac{\sqrt{2} M}{\frac{m}{C}
  \lambda_2 - 1} + \frac{\sqrt{2m} M}{\left(\frac{m}{C} \lambda_2 -
  1\right)^2}
\end{equation}
Note that this is the hypothesis stability of the algorithm. Let
$\mh_S (x_i)$ denote the predicted target value of the $i$th
vertex under $\mh_S$ (i.e.~the $i$th coordinate of
$\mh_S$). The cost-stability is given by:
\begin{equation}
\left| (\mh_S (x_i) - y_i)^2 - (\mh_{S'} (x_i) - y_i)^2 \right| \leq 4M
  \linf{\mh_S - \mh_{S'}}.
\end{equation}
Substituting the upper bound on $\linf{\mh_S - \mh_{S'}}$
derived in Equation~\ref{eqn:linf} into the above expression yields the
statement of the theorem.
\end{proof}
The generalization bound we just presented differs in several respects
from that of \cite{belkin04}. Our bound explicitly depends on both
$m$ and $u$ while theirs shows only a dependency on $m$. Also, our
bound does not depend on the number of times a point is sampled in the
training set (parameter $t$), thanks to our analysis based on sampling
without replacement.

Contrasting the stability coefficient of Belkin's algorithm with the
stability coefficient of \LTR\ (Theorem~\ref{thm:treg_stab}), we note
that it does not depend on $C'$ and $\bloc$. This is because unlabeled
points do not enter the objective function, and thus $C' \!=\! 0$ and
$\widetilde{y}(x) \!=\! 0$ for all $x \in X$. However, the stability
does depend on the second smallest eigenvalue $\lambda_2$ and the
bound diverges as $\lambda_2$ approaches $\frac{C}{m}$. Actually, the
bound in Theorem~\ref{thm:belkin-gen} will converge so long as
$\lambda_2 \!=\!  \Omega(1/m)$. As observed empirically by
\cite{Cortes&Mohri2006}, this algorithm does not perform as well in
comparison with \LTR.

\subsection{Cost stability of graph Laplacian regularization algorithm}
\label{sec:constrained_cost}

Here we give a cost-stability analysis of the graph Laplacian
regularization algorithm of \cite{belkin04}. To do so, we show that
the algorithm can in fact be viewed as a special instance of the
family of \LTR\ algorithms. Theorem~\ref{thm:treg_stab} can then be
applied in this instance with a bound on the cost stability
coefficient.

To show that that the graph Laplacian algorithm is a specific \LTR\
algorithm, we need to prove that the regularization term $\mh^\top \L
\mh$ corresponds to the square of a norm in some reproducing kernel
Hilbert space (RKHS). We show a more general result valid for all
positive semi-definite symmetric matrices $\L$. We denote by $\L^+$
the pseudo-inverse of a matrix $\L$.

\begin{theorem}
\label{th:rhks}
  Let $H_1$ be a vector space such that $H_1 \subseteq \range(\L)$,
  then the regularization term $\mh^\top \L \mh$ coincides with the
  square of the norm in the RKHS defined by the kernel matrix $\L^+$.
\end{theorem}
\begin{proof}
  We need to show that there exists a kernel $K$ such that $\mh^\top
  \L \mh = \norm{\mh}_K^2$ for all $h \in H_1$, where $\norm{\cdot}_K$
  is the norm in the RKHS associated to $K$.  This condition can be
  rewritten as $\mh^\top \L \mh = \iprod{\mh}{\mh}_K$, and more
  generally in terms of the inner product of $\mh, \mh' \in H_1$ as
\begin{equation}
{\mh'}^\top \L \mh = \iprod{\mh'}{\mh}_K.
\end{equation}
Let $\K$ denote the Gram matrix of $K$ for the sample $S$. Select
$\mh'$ to be $\K \me_i$, where $\me_i$ the $i$th unit vector of $H$.
Then, the equality is equivalent to  
\begin{equation}
  \forall i \in [1, m + u], \ \me_i^\top \K \L \mh = \iprod{\K \me_i}{\mh}_K = \iprod{K(x_i, \cdot)}{\mh}_K = h(x_i) = \me_i^\top \mh,
\end{equation}
where we used the reproducing property of the inner product. Since the
equality $\me_i^\top \K \L \mh = \me_i^\top \mh$ holds for all $i \in [1, m + u]$,
this is equivalent to the following,
\begin{equation}
 \forall \mh \in H_1, \ \K \L \mh = \mh .
\end{equation}
$\K = \L^+$ verifies this equality. Indeed, by the properties of the
pseudo-inverse, $\L^+ \L$ is the projection over $\range(\L)$. Since
by assumption $H_1 \subseteq \range(L)$, we can write $\L^+ \L \mh =
\mh$.
\end{proof}
In the particular case of the graph Laplacian, when the graph is connected
and the space $H_1$ orthogonal to $\mat 1$ coincides with $\range(\L)$ and
the result of the theorem holds.

\begin{corollary}
\label{cor:18}
  Any constraint optimization algorithm of the form (\ref{eqn:belkin})
  with $H_1 \!\subseteq\! \range(\L)$ is a special instance of the
  \LTR\ algorithms. In particular, the graph Laplacian regularization
  algorithm of \cite{belkin04} is a specific instance of the \LTR\
  algorithms.
\end{corollary}
The following theorem gives a bound on the cost stability of the graph
Laplacian algorithm.
\begin{theorem}
  \label{thm:belkin stability}
  Assume that the hypothesis set $H$ is bounded; that is, for all $h
  \in H$, and $x \in X$, $| h(x) - y(x) | \leq M$. Then, the graph
  Laplacian regularization algorithm of \cite{belkin04} has uniform
  stability $\beta$ with
  \begin{equation}
    \beta \leq \frac{4CM^2}{m} \min \Big\{ \frac{1}{\lambda_2},
    \rho_G \Big\},
  \end{equation}
  where $\lambda_2$ is the second smallest eigenvalue of the Laplacian
  matrix and $\rho_G$ the diameter of the graph $G$.
\end{theorem}
\begin{proof}
  By Corollary~\ref{cor:18}, the graph Laplacian regularization
  algorithm of \cite{belkin04} is a special case of the \LTR\
  algorithms. Thus, Theorem~\ref{thm:treg_stab} can be applied to
  determine its stability coefficient, with the term $AM$ bounding 
  $\abs{h(x) - y(x)}$ in that theorem replaced by $M$ here:
\begin{equation}
\label{belkin:stabcoeff}
  \beta \leq \frac{4CM^2 \kappa^2}{m}.
\end{equation}
Furthermore, using the same techniques as \citep{herbster05}, we can
bound $\mh^\top \L^+ \mh$ and thus $\kappa^2$ in terms of the second
smallest eigenvalue of the Laplacian matrix $\lambda_2$ and the
diameter of the graph $\rho_G$ as: $\kappa^2 \!\leq\! \min \big\{
\frac{1}{\lambda_2}, \rho_G \big\}$.  Substituting this upper bound in
Equation~\ref{belkin:stabcoeff} yields the statement of the theorem.
\end{proof}
The following theorem gives a novel stability generalization bound
for the algorithm of \cite{belkin04} in terms of the second eigenvalue
of the Laplacian and the diameter of the graph.

\begin{theorem}
  \label{thm:belkin}
  Let $H$ be a bounded hypothesis set. Let $G$ be a connected graph
  with diameter $\rho_G$ and $L$ be the associated Laplacian kernel with
  second smallest eigenvalue $\lambda_2$. Let $S$ be a random subset
  of labeled points of size $m$ drawn from the vertex set $X$. Let
  $\mh$ be the hypothesis returned by Equation~\ref{eqn:belkin} when
  trained on $X = (S, T)$. Then, for any $\e > 0$,
  \begin{equation}
    R(\mh) \leq \h R(\mh) + \frac{4CM^2 \kappa^2}{m} +
    \left( \frac{8CM^2 \kappa^2}{m}+ \frac{M^2(m + u)}{mu} \right)
    \sqrt{\frac{\ln(1/\delta) \alpha(m, u)}{2}},
  \end{equation}
  where $\alpha (m, u) = \frac{mu}{m + u - 1/2}$ and $\kappa^2 = \min \{
  1 / \lambda_2, \rho_G \}$.
\end{theorem}
\begin{proof}
  The result follows directly from Theorem~\ref{thm:generalization} and
  the stability coefficient $\beta$ derived in Theorem~\ref{thm:belkin
    stability}.
\end{proof}

\subsection{General case}
\label{sec:general_case}

The previous sections demonstrated the stability benefits of
constraints of the type $\u^\top \mh \!=\! 0$, which helped us bound
the stability of the graph Laplacian regularization algorithm and
derive stability-based generalization bounds.

This idea and in fact much of the results presented for this
particular algorithm can be generalized. To ensure stability, it
suffices that the optimization constraint restricts the hypothesis set
$H_1$ to be a subset of $\range(\L)$. By Theorem~\ref{th:rhks}, the
regularization term then corresponds to an RKHS norm
regularization. Orthogonality with respect to a single vector may not
be sufficient to ensure $H_1 \subseteq \range(\L)$. In fact, this does
not hold even for the graph Laplacian regularization algorithm if the
graph $G$ is not connected since the dimension of the null space of
$\L$ is then more than one. But, the constraints can be augmented to
guarantee this property by imposing orthogonality with respect to the
null space. More generally, one might wish to impose orthogonality with
respect to some space that guarantees that the smallest non-zero
eigenvalue over $H_1$ is not too small, for example by excluding
eigenvalue $\lambda_2$ if it is too small.

In particular, ``stable'' versions of the algorithms presented in
Section~\ref{sec:unconstrained} $\CM, \LLR$, and $\GMF$ can be
derived by augmenting their optimization problems with such
constraints.  Recall that the stability bound in
Proposition~\ref{prop:closed_form} is inversely proportional to the
smallest eigenvalue $\lambda_m (\mat{Q})$. The main difficulty with
using the proposition for these algorithms is that $\lambda_m
(\mat{Q}) = 0$ in each case. Let $\mat{v}_m$ denote the eigenvector
corresponding to $\lambda_m (\mat{Q})$ and let $\lambda_2$ be the
second smallest eigenvalue of $\mathbf{Q}$. One can modify
(\ref{eqn:obj_fun}) and constrain the solution to be orthogonal to
$\mat{v}_m$ by imposing $\mh \cdot \mat{v}_m = 0$. In the case of
\cite{belkin04}, $\mat{v}_m = \mat{1}$. This modification, motivated
by the algorithm of~\cite{belkin04}, is equivalent to increasing the
smallest eigenvalue to be $\lambda_2$.

As an example, by imposing the additional constraint, we can show
that the stability coefficient of $\CM$  becomes bounded by
$O(C/\lambda_2)$, instead of $\Theta(1)$. Thus, if $C = O(1/m)$
and $\lambda_2 = \Omega(1)$, it is bounded by $O(1/m)$ and the
generalization bound converges as $O(1/m)$.

\section{Experiments}
\label{sec:experiments}

This section reports the results of experiments using our
stability-based generalization bound for model selection for the \LTR\
algorithm. A crucial parameter of this algorithm is the stability
coefficient $\bloc(r)$ of the local algorithm, which computes
pseudo-targets $\widetilde{y}_{x}$ based on a ball of radius $r$
around each point.  We derive an expression for $\bloc(r)$ and show,
using extensive experiments with multiple data sets, that the value
$r^*$ minimizing the bound is a remarkably good estimate of the best $r$ for the 
test error. This demonstrates the benefit of our generalization bound
for model selection, avoiding the need for a held-out validation
set.

\begin{figure*}[t]
\begin{center}
\begin{tabular*}{.9\textwidth}{@{\extracolsep{\fill}}ccc}
\rotatebox{-90}{\ipsfig{.16}{figure=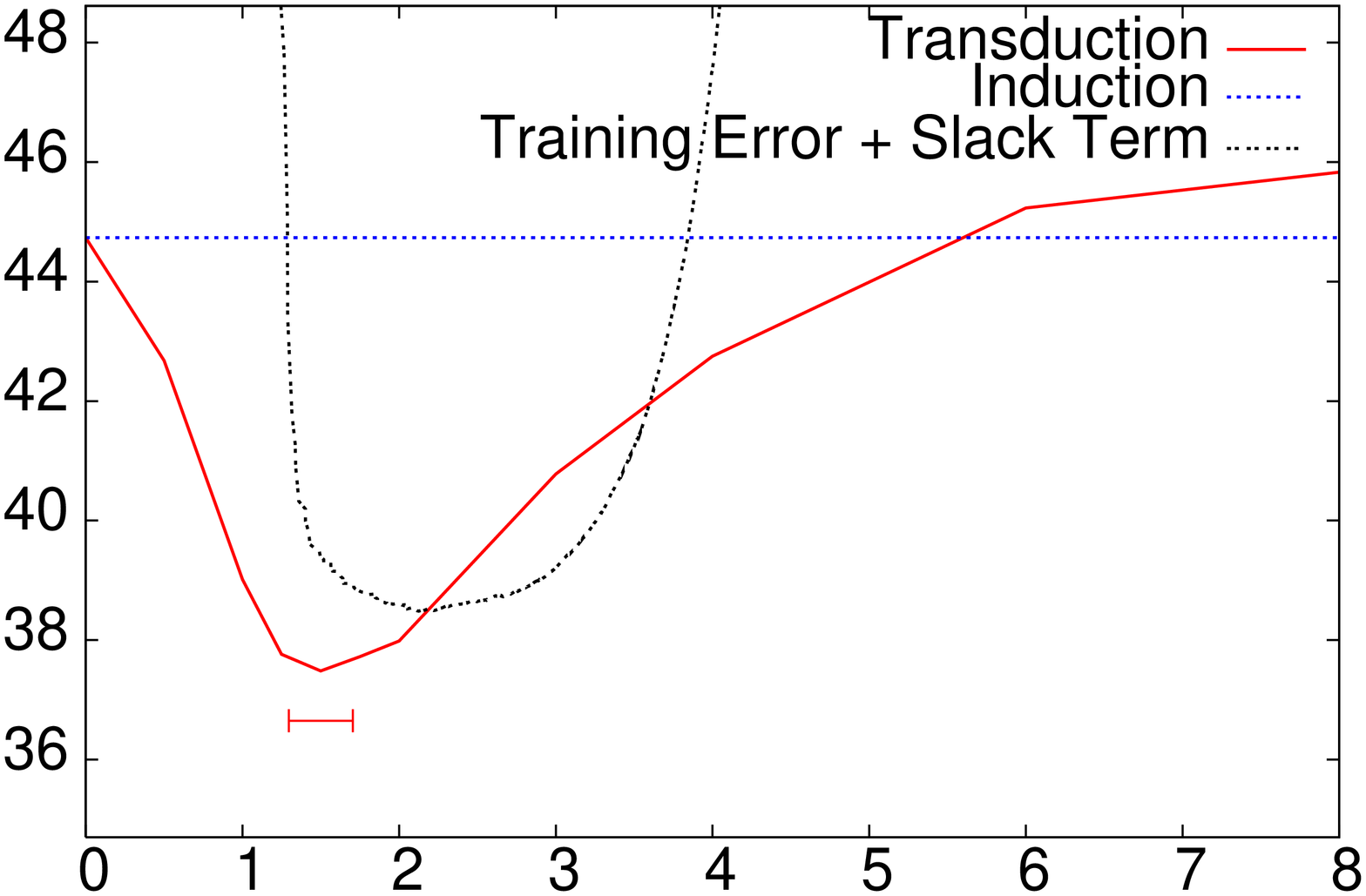}} & \rotatebox{-90}{\ipsfig{.16}{figure=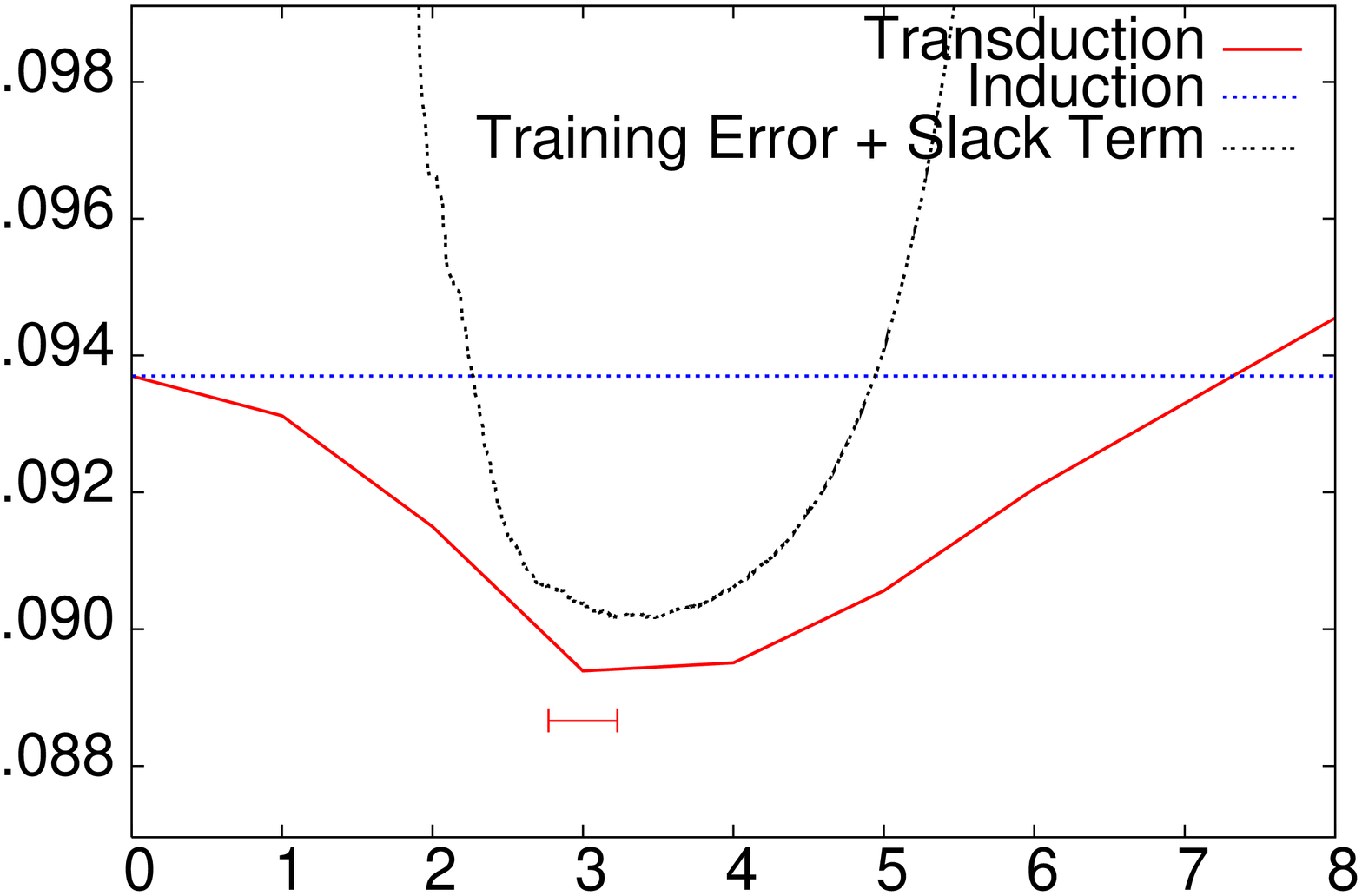}} & \rotatebox{-90}{\ipsfig{.16}{figure=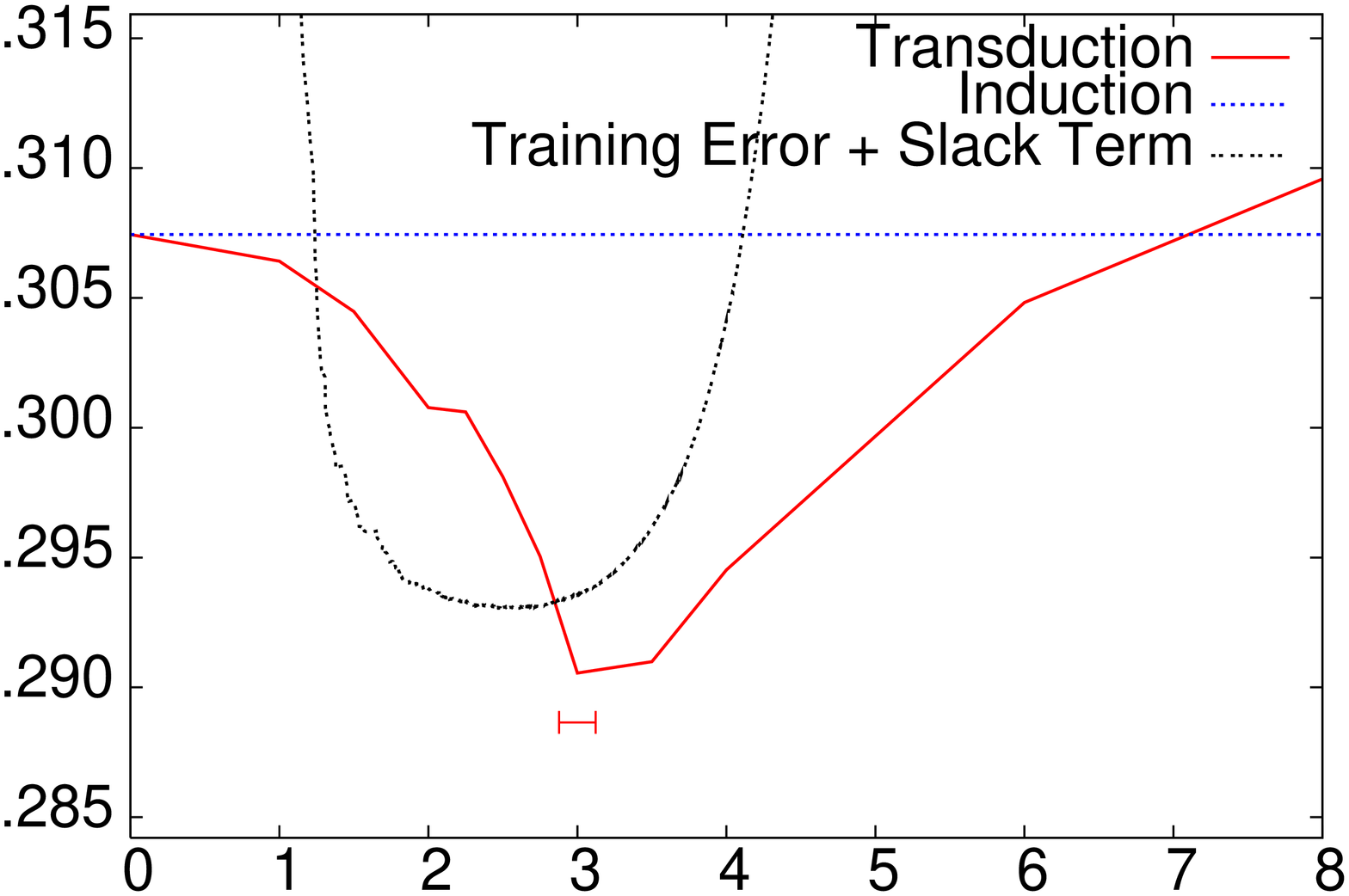}}\\
(a) & (b) & (c)
\end{tabular*}
\end{center}
\vskip -.2in
\caption{MSE against the radius $r$ of \LTR\ for three data sets: (a)
  Boston Housing. (b) Ailerons. (c) Elevators. The small horizontal
  bar indicates the location (mean $\pm$ one standard deviation) of
  the minimum of the empirically determined $r$.}
\label{fig:MSEs}
\vskip -.15in
\end{figure*}
The experiments were carried out on several publicly available
regression data sets: {\em Boston Housing}, {\em Elevators} and {\em
  Ailerons}\footnote{{\tt 
    www.liaad.up.pt/\~{}ltorgo/Regression/DataSets.html.}}. For each
of these data sets, we used $m=u$, inspired by the observation that,
all other parameters being fixed, the bound of
Theorem~\ref{thm:generalization} is tightest when $m = u$. The value
of the input variables were normalized to have mean zero and variance
one. For the Boston Housing data set, the total number of examples was
506. For the Elevators and the Ailerons data set, a random subset of
2000 examples was used. For both of these data sets, other random
subsets of 2000 samples led to similar results. The Boston Housing
experiments were repeated for 50 random partitions, while for the
Elevators and the Ailerons data set, the experiments were repeated for
20 random partitions each. Since the target values for the Elevators
and the Ailerons data set were extremely small, they were scaled by a
factor 1000 and 100 respectively in a pre-processing step.

In our experiments, we estimated the pseudo-target of a point $x' \in
T$ as a weighted average of the labeled points $x \in N(x')$ in a
neighborhood of $x'$. Thus, $\label{eq:y_est}\widetilde{y}_{x'} =
{\sum_{x \in N(x')} \alpha_x y(x)}/{\sum_{x \in N(x')} \alpha_x}$.  We
considered two weighting approaches, as discussed in
\citep{Cortes&Mohri2006}, defining them in terms of the inverse of the
distance between $\Phi(x)$ and $\Phi(x')$ (i.e.~$\alpha_x = (1 + |\!|
\Phi(x) - \Phi(x') |\!|)^{-1}$), and in terms of a similarity measure
$K(x, x')$ captured by a kernel $K$ (i.e.~$\alpha_x = K(x, x')$).  In
our experiments, the two approaches produced similar results. We
report the results of kernelized weighted average with a Gaussian
kernel.
\begin{lemma}
\label{lemma:bloc_bound}
Let $r \geq 0$ be the radius of the ball around an unlabeled point $x'
\in X$ that determines the neighborhood $N(x')$ of $x'$ and let $m(r)$
be the number of labeled points in $N(x')$. Furthermore, assume that
the values of the labels are bounded (i.e.~for all $x \in X, |y(x)|
\leq M$ for some $M > 0$ and that all the weights in
(\ref{eq:y_est}) are non-negative (i.e.~for all $x, \alpha_x \geq
0$). Then, the stability coefficient of the weighted average algorithm
for determining the estimate of the unlabeled point $x'$ is bounded
by:
\begin{equation}
\bloc \leq \frac{4\alpha_{\max}M}{\alpha_{\min}m(r)},
\end{equation}
where $\alpha_{\max} = \max_{x \in N(x')} \alpha_x$ and $\alpha_{\min}
= \min_{x \in N(x')} \alpha_x$. 
\end{lemma}
\begin{proof}
  We consider the change in the estimate as a point is removed from
  $N(x')$ and show that this is at most
  $\frac{2\alpha_{\max}M}{\alpha_{\min} m(r)}$. The statement of the
  lemma then follows straightforwardly from the observation that
  changing one point is equivalent to removing one point and adding
  another point.

  Let $N(x') = \{ x_1, \ldots, x_{m(r)} \}$. For ease of notation,
  assume that $n = m(r)$. Consider the effect of removing $x_n$ from
  the neighborhood $N(x')$. The estimate changes by:
  \begin{equation*}
    \frac{\sum_{i=1}^n \alpha_i y_i}{\sum_{i=1}^n \alpha_i} -
    \frac{\sum_{i=1}^{n-1} \alpha_i y_i}{\sum_{i=1}^{n-1} \alpha_i}.
  \end{equation*}
  Thus, the stability $\bloc$ can be bounded as follows:
  \begin{align*}
    \bloc & \leq \left|\frac{\sum_{i=1}^n \alpha_i y_i}{\sum_{i=1}^n
      \alpha_i} - \frac{\sum_{i=1}^{n-1} \alpha_i
      y_i}{\sum_{i=1}^{n-1} \alpha_i} \right| \\ 
    & \leq \frac{\alpha_n | y_n |}{\sum_{i=1}^n \alpha_i} +
    \sum_{i=1}^{n-1} \alpha_i |y_i| \left( \frac{1}{\sum_{i=1}^{n-1}
      \alpha_i} - \frac{1}{\sum_{i=1}^n \alpha_i} \right) \\ 
    & = \frac{\alpha_n |y_n|}{\sum_{i=1}^n \alpha_i} +
    \frac{\sum_{i=1}^{n-1} \alpha_i |y_i|}{\sum_{i=1}^{n-1} \alpha_i}
    \cdot \frac{\alpha_n}{\sum_{i=1}^{n} \alpha_i} \\
    & \leq \frac{2\alpha_n M}{\sum_{i=1}^n \alpha_i} \leq \frac{2
      \alpha_{\max} M}{\alpha_{\min} n} \leq \frac{2 \alpha_{\max} M}{\alpha_{\min} m(r)},
  \end{align*}
which proves the statement of the lemma.
\end{proof}
\begin{corollary}
Using the notation of Lemma~\ref{lemma:bloc_bound}, the stability
coefficient of the kernelized weighted average algorithm with a
Gaussian kernel $K$ with parameter $\sigma$ is bounded by:
\begin{equation*}
\bloc \leq \frac{4M}{m(r) e^{-2r^2 / \sigma^2}}. 
\end{equation*}
\end{corollary}
\begin{proof}
  This follows directly from Lemma~\ref{lemma:bloc_bound} using the
  observation that for a Gaussian kernel $K$, $K(x, x') \leq 1$, and
  for $x, x'$ such that $|\!| x |\!| \leq r$ and $|\!| x' |\!| \leq
  r$, $|\!| x - x' |\!| \leq 2r$. Thus, $K(x, x') \geq e^{-2r^2 /
    \sigma^2}$. 
\end{proof}
\begin{corollary}
Using the notation of Lemma~\ref{lemma:bloc_bound}, the stability
coefficient of the weighted average algorithm, where weights are
determined by the inverse of the distance in the feature space,
i.e.~$\alpha_x = (1 + |\!| \Phi(x) - \Phi(x') |\!|)^{-1}$ is bounded
by:\footnote{1 is added to the weight to make the weights between 0
and 1.}
\begin{equation*}
  \bloc \leq \frac{(2r + 1)2M}{m(r)}.
\end{equation*}
\end{corollary}
\begin{proof}
  This follows directly from Lemma~\ref{lemma:bloc_bound} using the
  observation that for all $x \in N(x')$, 
\begin{equation*}
  0 \leq |\!| \Phi(x) - \Phi(x') |\!| \leq 2r. 
\end{equation*}
\end{proof}
To estimate $\bloc$, one needs an estimate of $m(r)$, the number of
examples in a ball of radius $r$ from an unlabeled point $x'$. In our
experiments, we estimated $m(r)$ as the number of labeled examples in a ball of
radius $r$ from the origin. Since all features are normalized to mean
zero and variance one, the origin is also the centroid of the set $X$.

We implemented a dual solution of \LTR\ and used Gaussian kernels, for
which, the parameter $\sigma$ was selected using cross-validation on
the training set. Experiments were repeated across 36 different pairs
of values of $(C, C')$. For each pair, we varied the radius $r$ of the
neighborhood used to determine estimates from zero to the radius of
the ball containing all points.

Figure~\ref{fig:MSEs}(a) shows the mean values of the test MSE of our
experiments on the Boston Housing data set for typical values of $C$
and $C'$.  Figures~\ref{fig:MSEs}(b)-(c) show similar results for the
Ailerons and Elevators data sets. For the sake of comparison, we also
report results for induction. The induction algorithm we chose was
Kernel Ridge Regression (since it is analogous to \LTR with the choice
of $C' = 0$). The relative standard deviations on the MSE are not
indicated, but were typically of the order of 10\%. \LTR\ generally
achieves a significant improvement over induction.

The generalization bound we derived in
Equation~\ref{eqn:main_stability_bound} consists of the training error and
a complexity term that depends on the parameters of the \LTR\
algorithm ($C, C', M, m, u, \kappa, \bloc, \delta$). Only two terms
depend upon the choice of the radius $r$: $\h R(h)$ and
$\bloc$.  Thus, keeping all other parameters fixed, the theoretically
optimal radius $r^*$ is the one that minimizes the training error plus
the slack term. The figures also include plots of the training error
combined with the complexity term, appropriately scaled. The empirical
minimization of the radius $r$ coincides with or is close to
$r^*$. The optimal $r$ based on test MSE is indicated with error bars.

\section{Conclusion}

We presented a comprehensive analysis of the stability of transductive
regression algorithms with novel generalization bounds for a number of
algorithms. Since they are algorithm-dependent, our bounds are often
tighter than those based on complexity measures such as the
VC-dimension.  Our experiments also show the effectiveness of our
bounds for model selection and the good performance of \LTR\
algorithms in practice. Our analysis can also guide the design of
algorithms with better stability properties and thus generalization
guarantees, as discussed in Section~\ref{sec:general_case}. The general
concentration bound for uniform sampling without replacement proved
here can be of independent interest in a variety of other machine 
learning and algorithmic analyses.

\bibliography{jstr}
\end{document}